\documentclass[review]{elsarticle}
\graphicspath{ {./figures/} }
\usepackage{hyperref}
\usepackage{float}
\usepackage{verbatim} %comments
\usepackage{apalike}
\restylefloat{figure}
\restylefloat{table}
\usepackage{graphicx}
\usepackage{tabularx, url, array, multirow}
\usepackage{tablefootnote}
\usepackage{xcolor}
\usepackage{cleveref}
\crefname{section}{§}{§§}
\Crefname{section}{§}{§§}
\usepackage{amssymb}
\usepackage{eucal}

\usepackage{framed}
\definecolor{shadecolor}{rgb}{0.9,0.9,0.9}

\journal{Expert Systems with Applications}

\biboptions{authoryear}

\begin{document}
\begin{frontmatter}

\begin{titlepage}
\begin{center}
\vspace*{1cm}

\textbf{ \large Crude Oil-related Events Extraction and Processing: A Transfer Learning Approach}

\vspace{1.5cm}

% Author names and affiliations
Meisin Lee$^{a}$ (mei.lee@monash.edu), Lay-Ki Soon$^a$ (soon.layki@monash.edu), Eu-Gene Siew$^a$ (siew.eu-gene@monash.edu) \\

\hspace{10pt}

\begin{flushleft}
\small  
$^a$ Monash University (Malaysia Campus), Jalan Lagoon Selatan, Bandar Sunway, 47500, Selangor, Malaysia

\begin{comment}
Clearly indicate who will handle correspondence at all stages of refereeing and publication, also post-publication. Ensure that phone numbers (with country and area code) are provided in addition to the e-mail address and the complete postal address. Contact details must be kept up to date by the corresponding author.
\end{comment}

\vspace{1cm}
\textbf{Corresponding Author:} \\
Meisin Lee \\
%Full address of the corresponding author, including the country name \\
Tel: +603-5514 6000 \\
Email: mei.lee@monash.edu

\end{flushleft}        
\end{center}
\end{titlepage}

\title{Crude Oil-related Events Extraction and Processing: A Transfer
Learning Approach}

\author[label1]{Meisin Lee\corref{cor1}}
\ead{mei.lee@monash.edu}

\author[label1]{Lay-Ki Soon}
\ead{soon.layki@monash.edu}

\author[label1]{Eu-Gene Siew}
\ead{siew.eu-gene@monash.edu}

\cortext[cor1]{Corresponding author.}
\address[label1]{Monash University (Malaysia Campus), Jalan Lagoon Selatan, Bandar Sunway, 47500, Selangor, Malaysia}

\begin{abstract}
One of the challenges in event extraction via traditional supervised learning paradigm is the need for a sizeable annotated dataset to achieve satisfactory model performance. It is even more challenging when it comes to event extraction in the finance and economics domain, a domain with considerably fewer resources. This paper presents a complete framework for extracting and processing crude oil-related events found in \textit{CrudeOilNews} corpus \cite{lee2022crudeoilnews}, addressing the issue of annotation scarcity and class imbalance by leveraging on the effectiveness of transfer learning. Apart from event extraction, we place special emphasis on event properties (Polarity, Modality, and Intensity) classification to determine the factual certainty of each event. %Through this we hope to contribute towards a richer understanding and more accurate representation of events grounded in accurate arguments extraction as well as accurate classification of event properties. 
We build baseline models first by supervised learning and then exploit Transfer Learning methods to boost event extraction model performance despite the limited amount of annotated data and severe class imbalance. This is done %through improving model representation 
via methods within the transfer learning framework such as \textit{Domain Adaptive Pre-training, Multi-task Learning} and \textit{Sequential Transfer Learning.} Based on experiment results, we are able to improve all event extraction sub-task models both in F1 and MCC\footnote{Matthews correlation coefficient}-score as compared to baseline models trained via the standard supervised learning. Accurate and holistic event extraction from crude oil news is very useful for downstream tasks such as understanding event chains and learning event-event relations, which can be used for other downstream tasks such as commodity price prediction, summarisation, etc. to support a wide range of business decision making.

%understanding the correlation of current world affairs and crude oil price movement, which can then be used to %Events found in Commodity News can be categorized into Macro-Economic, Geo-political and supply-demand related, they fall under the Finance and Economics domain.  The event extraction task is broken down into the following sub-tasks: (1) Entity Mention Extraction, (2) Event Extraction and (3) Event Properties classification
\end{abstract}

\begin{keyword}
event extraction \sep transfer learning \sep commodity news \sep information extraction \sep low resource domain
\end{keyword}

\end{frontmatter}

\section{Introduction} \label{sec:intro}
Event extraction is an important task in Information Extraction.  It is the process of gathering knowledge about incidents found in texts, automatically identifying information about what happened, when it happened, and other details. Event extraction has long been a challenging task, addressed mostly with supervised methods\footnote{Apart from supervised methods, there is a smaller number of work proposed the use some form of Weak Supervision, Distant Supervision, etc (see \cite{8918013} for a survey of existing event extraction methods).} that require massive amounts of annotated data. However, annotated data is hard and expensive to obtain. This is evident that even canonical datasets such as ACE2005 and TAC KBP are moderate in size. This challenge is even more apparent in specialised domains such as finance and economics, where only experts can provide reliable labels \cite{konyushkova2017learning}. The challenge of annotation scarcity and class imbalance is acknowledged in \cite{chen2021}, to which the authors proposed to use transfer learning by using a number of corpora within the BioMedical domain to increase the coverage of event detection (event trigger detection) on their target corpus of the same domain.

In this work, we investigate the task of event extraction in \textit{CrudeOilNews} corpus, a dataset released by \cite{lee2022crudeoilnews} where commodity news articles are annotated for the task of event extraction. More information about the \textit{CrudeOilNews} corpus is laid out in Section \ref{sec:dataset}. Events found in commodity news articles are mainly Geo-political, Macro-economics, supply and demand in nature.
%which fall into the Finance and Economic domain. The main focus of this work is to build machine learning models to extract events from crude oil news. 
Transfer Learning has been proven to be effective for a wide range of applications, especially for low-resourced domains \cite{meftah2021hidden}. Inspired by this, we explore the usage of transfer learning to produce event extraction and event property classification models with the best possible accuracy despite of the limited training size. Transfer learning is a set of methods that leverages resources from other domains or resources intended from other tasks to train a model with better generalization properties. Resources from other domains are known as \textit{source domain}, while resources intended for a different task is known as \textit{source task}. %The knowledge gained in training a different source task or in training the same task in a different domain then applied to the \textit{target task} or \textit{target domain} \cite{ruder2019neural}. 
Transfer learning aims at performing a task on a target dataset using features learned from a source dataset \cite{pan2010survey}.
In this work, we experimented with various approaches within the Transfer Learning  paradigm\footnote{The definition of Transfer Learning is aligned to the taxonomy of Transfer Learning described in \cite{ruder2019neural}.}, namely \textit{Domain Adaptive Pre-training, Sequential Transfer Learning} and \textit{Multi-task Learning}. Definition of the various types of Transfer Learning are laid out in Section \ref{subsubsec:taxonomy}. 

\subsection{Dataset} \label{sec:dataset}
%Constructing supervised training data is prohibitively expensive as this requires the use of expert knowledge in macro-economic, business and finance field.herefore, due to this limitation of having a limited amount of labeled training data, most text mining models cannot directly utilize supervised training approach.  To the best of our knowledge, we found only one dataset annotated for the task of event extraction for Geo-political and Macro-economics events. This further supports our observation that resources in this domain are somewhat limited. 
The dataset used here is the \textit{CrudeOilNews} corpus introduced in \cite{lee2022crudeoilnews}. The dataset size is moderate where it contains, in total,  \textbf{425 documents}, \textbf{approx. 11k events}, about \textbf{approx. 23k arguments}, and each event is classified according to Polarity, Modality, and Intensity properties. There are 21 entity types\footnote{for simplicity and convenience, \textbf{values} and \textbf{temporal expressions} are also considered as entity mentions}, 18 event types. Broadly, the events can be grouped into the following main categories:

\begin{itemize}
\setlength\itemsep{-0.3em}
    \item \textbf{geo-political:} Geo-political tension, civil unrest, embargo/sanctions, trade tensions, and other forms of geo-political crisis.
    \item \textbf{macro-economic:} US employment data, economic/gross domestic product (GDP) growth, economic outlook, growth forecast, supply and demand
    \item \textbf{Supply-Demand related}: oversupply
    \item \textbf{price movement:} price increase, price decrease, price forecast. 
\end{itemize} 
As for event arguments, there are 21 argument types; each event type is associated with a set of argument roles. For more details, such as entity mention examples and description of event types, refer to \ref{app_dataset}.

%Based on the list above, we proposed a solution to address the first two challenges. The proposed solution consist of leveraging transfer learning to overcome the issue of limited labeled data and class imbalance. As for last two characteristics, 

\subsection{Definitions} \label{sec:definitions}
Before we dive into the technical details of the proposed solution, this section is dedicated to laying out the terminologies and task descriptions. 
\paragraph{Terminologies:}
\begin{enumerate}
    \item An \textbf{entity mention} is an explicit mention of an entity in a text that has an entity type.
    \item An \textbf{event trigger} is the main word(s) that most clearly expresses the occurrence of an event, usually a word or a multiworld phrase. It can come in the form of verb, noun, adjective, or adverb). 
    \item An \textbf{event argument} is an argument filler that plays a certain role in an event.
    \item \textbf{Polarity}, which is also known as negation in \cite{morante2012sem}, denotes whether an event actually happened or was negated, not to be confused with the polarity in sentiment analysis. Its value can be {\fontfamily{qcr}\selectfont POSITIVE} or {\fontfamily{qcr}\selectfont NEGATIVE}.
    \item \textbf{Modality}, also known as hedge in \cite{farkas2010conll}, denotes whether an event actually happened or will happen in the future, or whether it is a generic event. Its value can be {\fontfamily{qcr}\selectfont ASSERTED} or {\fontfamily{qcr}\selectfont OTHER}.
    \item \textbf{Intensity} denotes if an event further intensified and lessen. Its value can be {\fontfamily{qcr}\selectfont INTENSIFIED}, {\fontfamily{qcr}\selectfont NEUTRAL}, and {\fontfamily{qcr}\selectfont EASED}. 
    %\item \textbf{Transfer Learning}: Definition of Transfer Learning techniques mentioned here follows the taxonomy by \cite{ruder2019neural} shown in Figure \ref{fig:taxonomy} in the Appendix section . 
    %\item \textbf{Actie Learning}...
\end{enumerate}
According to \cite{lee2022crudeoilnews}, the definition of Event Polarity and Modality in \textit{CrudeOilNews} corpus are aligned with \textit{ACE2005}, while Event Intensity is a newly defined property specially crafted for events found in commodity news.

\begin{figure}[ht]
\centering
    \includegraphics[width=0.75\textwidth]{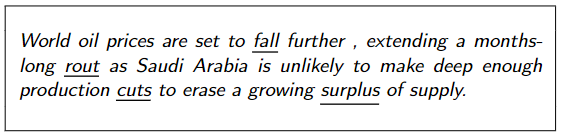}
    \caption{This is an example taken from \textit{CrudeOilNews} corpus; event trigger words are underlined.}
    \label{fig:example1}
    \vspace{-0.5em}
\end{figure}

We propose a solution for sentence-level Event Extraction, which is made up of a few sub-tasks. These tasks\footnote{We follow the same naming convention in \cite{nguyen2019one}} are described below:

\paragraph{Tasks}
\begin{enumerate}
    \item Entity Mention Detection \textbf{(EMD)}: a task to detect entity mentions (named or nominal) and assign each token an entity type or NONE for tokens that are not an entity mention. See Table \ref{table:task_EMD_ED} top section for EMD task on the example shown in Figure \ref{fig:example1}.
    \item Event Extraction : 
        \begin{enumerate}
            \item Event Detection \textbf{(ED)}: similar to EMD, it is a task to detect event trigger word(s) and assign it to an event type or NONE for tokens that are not an event trigger. See Table \ref{table:task_EMD_ED} bottom section for ED task on the example in Figure \ref{fig:example1}.
            \item Argument Role Prediction \textbf{(ARP)}: a task aims to assign an argument role label or NONE to a candidate entity mention. If the candidate entity is not linked to the event. See Table \ref{table:task_ARP} for ARP task on the example shown in Figure \ref{fig:example1}.
        \end{enumerate}
    \item Event Properties Classification: Classifying each event's in terms of their Polarity, Modality, and Intensity classes. See Table \ref{table:example} for the classification of each event against Polarity, Modality and Intensity. 
\end{enumerate}

\begin{table}[h!]   
    \caption{Entity Mention Detection (EMD) and Event Detection (ED) for the example sentence shown in Figure \ref{fig:example1}.}
    \centering \small
    \begin{tabular}{ | p{0.13\linewidth} | p{0.3\linewidth} | p{0.45\linewidth} |}  \hline
    \textbf{Task } & \textbf{Text} & \textbf{Entity Type} \\ \hline
    \multirow{7}{*}{\textbf{EMD}} & World & Location \\ \cline{2-3}
     & oil & {\fontfamily{qcr}\selectfont COMMODITY} \\ \cline{2-3}
     & prices & {\fontfamily{qcr}\selectfont FINANCIAL ATTRIBUTE} \\ \cline{2-3}
     & months-long & {\fontfamily{qcr}\selectfont DATE} \\ \cline{2-3}
     & Saudi Arabia & COUNTRY \\ \cline{2-3}
     & production & {\fontfamily{qcr}\selectfont FINANCIAL ATTRIBUTE}  \\ \cline{2-3}
     & supply & {\fontfamily{qcr}\selectfont FINANCIAL ATTRIBUTE} \\ \hline \hline
     \textbf{Task} & \textbf{Trigger Word(s)} & \textbf{Event Type} \\ \hline
    \multirow{4}{*}{\textbf{ED}} & fall & {\fontfamily{qcr}\selectfont MOVEMENT\_DOWN\_LOSS} \\ \cline{2-3}
     & rout & {\fontfamily{qcr}\selectfont SLOW-WEAK} \\ \cline{2-3}
     & cuts & {\fontfamily{qcr}\selectfont CAUSE} {\fontfamily{qcr}\selectfont MOVEMENT\_DOWN\_LOSS} \\ \cline{2-3}
     & surplus & {\fontfamily{qcr}\selectfont OVERSUPPLY} \\ \hline
      \end{tabular}
      \vspace{-1em}
    \label{table:task_EMD_ED}
\end{table}

\begin{table}[h!]   
    \caption{Argument Role Prediction (ARP) for the example sentence shown in Figure \ref{fig:example1}.}
    \centering \small
    \begin{tabular}{ |p{0.20\linewidth} | p{0.30\linewidth} | p{0.25\linewidth} |} \hline
    \textbf{Event} & \textbf{Text} & \textbf{Argument Role}\\ \hline
    \underline{fall} & oil & {\fontfamily{qcr}\selectfont ITEM} \\ \cline{2-3}
     & prices & {\fontfamily{qcr}\selectfont ATTRIBUTE}\\ \hline
    \underline{rout} & months-long & {\fontfamily{qcr}\selectfont DURATION}\\ \hline
    \underline{cuts} & Saudi Arabia & {\fontfamily{qcr}\selectfont SUPPLIER}\\ \cline{2-3}
    & production & {\fontfamily{qcr}\selectfont ATTRIBUTE} \\ \hline
    \underline{surplus} & supply & {\fontfamily{qcr}\selectfont ATTRIBUTE}\\ \hline
      \end{tabular}
      \vspace{-1em}
    \label{table:task_ARP}
\end{table}

\begin{table}[h!] 
\caption{Event properties (Polarity, Modality and Intensity) for each of the events from the example sentence in Figure \ref{fig:example1}.} \label{table:example}
\begin{tabular}{ |l | l l l |}  \hline
\textbf{Event} & \textbf{Polarity} & \textbf{Modality} & \textbf{Intensity}\\ \hline
prices \underline{fall} & {\fontfamily{qcr}\selectfont POSITIVE} & {\fontfamily{qcr}\selectfont OTHER} & {\fontfamily{qcr}\selectfont NEUTRAL} \\
\underline{rout} & {\fontfamily{qcr}\selectfont POSITIVE} & {\fontfamily{qcr}\selectfont ASSERTED} & {\fontfamily{qcr}\selectfont INTENSIFIED} \\
production \underline{cuts} & {\fontfamily{qcr}\selectfont NEGATIVE} & {\fontfamily{qcr}\selectfont OTHER} & {\fontfamily{qcr}\selectfont NEUTRAL} \\
supply \underline{surplus} & {\fontfamily{qcr}\selectfont POSITIVE} & {\fontfamily{qcr}\selectfont OTHER} & {\fontfamily{qcr}\selectfont EASED} \\ \hline
\end{tabular}
\end{table}

Based on the example in Figure \ref{fig:example1} and information extracted via regular event extraction, which is made up subtasks (1) EMD and ED (Table \ref{table:task_EMD_ED}), and (2) APR (Table \ref{table:task_ARP}), we argue that the extracted information is incomplete without without taking into consideration its events' properties shown in Table \ref{table:example}, 

\subsection{Transfer Learning} \label{subsubsec:taxonomy}
\begin{figure}[h]
\centering
    \includegraphics[width=0.85\textwidth]{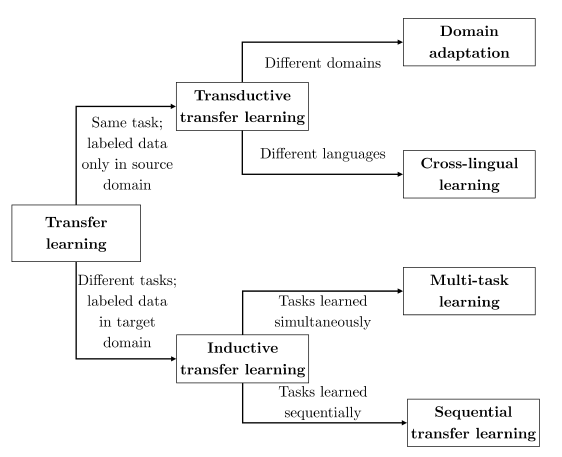}
    \caption{A taxonomy of Transfer Learning for NLP (image taken from \cite{ruder2019neural})}
    \label{fig:taxonomy}
\end{figure}

%We define a domain $D$ as a tuple ($X, P(X)$) where $X$ is the feature space and $P(X)$ is the marginal probability of the feature space. Next we define a task $T$ as a tuple ($y, P(y|x)$) where $y$ are the labels and $P(y|x)$ is the conditional distribution that our machine learning models try to learn. For each feature $x \in X$ we associate a label $y$. The task is associated with an objective function $P(y|x)$ which is optimized to learn the labels of the documents given the feature vector for each document. 
Here we provide the formal definition of Transfer Learning based on \cite{weiss2016survey, pan2010survey, alyafeai2020survey}. Given a source-domain task tuple ($D_s, T_s$) and different target domain-task pair($D_t, T_t$), we define transfer learning as the process of using the source domain and task in the learning process of the target domain task. 

The definition of transfer learning used here is aligned to \cite{ruder2019neural}, and the taxonomy diagram used in \cite{ruder2019neural} is shown in Figure \ref{fig:taxonomy}. According to this taxonomy, the main branches of Transfer Learning are:
\begin{enumerate}
    %\item Domain Adaptation ($D_s \neq D_t, T_s = T_t$): the process of adapting to a new domain (target domain has different data distribution from source domain). 
    \item \textbf{Transductive transfer Learning}: It is the setting where source and target tasks are the same; we don’t have labeled data at all or we have very few labeled data in the target domain, but sufficient labeled data in the source domain.
    \item \textbf{Inductive transfer learning}: It is the setting where the source task and the target task are different; labeled data is only available in the target domain.
\end{enumerate}

Based on the above definition and given that we intend to fully utilize the labels in our dataset (target domain) to train (or fine-tune) the model, we will only focus on \textbf{Inductive transfer learning} and its sub-types, namely:
\begin{enumerate}
    \item \textbf{Sequential Transfer Learning (STL)}: the process of learning multiple tasks ($T_1, T_2, ...., T_n$). At each step $t$, we learn a specific task $T_t$. %According to \cite{alyafeai2020survey}, this can be broken down into four categories, namely \textit{Fine-Tuning, Adapter Modules, Feature based and zero-shot.} 
    There are two types of STL, each illustrated by an example below: 
    \begin{enumerate}
        \item Cross-domain STL: A model is first trained on task $T$ using a source dataset $D_s$ and is then transferred to train on the target dataset $D_t$ on the same task ($T_s$ = $T_t$). In \cite{gui2018transferring}, authors used cross-domain STL to transfer a POS-tagging model trained using News, a resource-rich domain, to train on the same task using Tweets, a lower resource domain;
        \item Cross-task STL: A model is first trained on task $T_s$ and is then transferred to train on a different but related task $T_t$ in the same domain. This is seen in \cite{meftah2018neural}, where the authors first train a Named Entity Recognition (NER) model and then transfer the model to train on POS-tagging task in the same dataset \cite{meftah2018neural}.
        %Pre-traing language models are fine-tuned (or STL) to do other task such as Text Classification and shows SOTA results. 
    \end{enumerate}
    \item \textbf{Multi-task Learning (MTL)}: the process of learning multiple tasks ($T_1, T2,...,T_n$) at the same time. All tasks are learned in a parallel fashion.
    %\item Cross-lingual learning: the process of adapting to a different language in the target domain. This is excluded in this work because we work with English corpora only. 
    For example, both chunking and POS-tagging are trained concurrently in \cite{sogaard2016deep, ruder2017sluice}, as chunking has been shown to benefit from being jointly trained with low-level tasks such as POS tagging.
\end{enumerate}

\textbf{Negative Transfer} There are cases when transfer learning can lead to a drop in performance instead of improving it. Negative transfer refers to scenarios where the transfer of knowledge from the source to the target does not lead to improvements, but rather causes a drop in the overall performance of the target task that might be lower than that with a solely supervised training on in-target data \cite{torrey2010transfer}. There can be various reasons or various situation that resulted in negative transfer, such as:
\begin{enumerate}
    \item in MTL or STL task-centered transfer learning when the source task is not sufficiently related to the target task or if the transfer method could not leverage the relationship between the source and target tasks very well \cite{Rosenstein05totransfer};
    \item in domain adaptation when the source domain is dissimilar or less related to target domain \cite{meftah2021hidden, blitzer2007biographies, ruder2019neural, gui2018negative}
\end{enumerate}

%\subsection{Event Properties Classification}
%Typically Event Extraction involves only the first three sub-tasks: Entity Mention Detection (\textbf{EMD}), Event Detection (\textbf{ED}) and Argument Role Prediction (\textbf{ARP}). This work, on the other hand, aims to investigate Event Extraction and processing in a unified fashion, not just event extraction but also Event Properties classification - \textit{Polarity}, \textit{Modality} and \textit{Intensity}. As acknowledged by \cite{sauri2009factbank}, many computational linguistics points out the importance for systems to be sensitive to the veracity or factual certainty of events; that is to recognize whether events are presented as corresponding to actual situations in the world, situations that have not happened, or situations of uncertain interpretation. The examples below illustrate the importance of identifying event properties: As clearly shown, the events \underline{fall}, \underline{cuts}, and \underline{surplus} are speculated events (as opposed to an actual event that has taken place), on top of that \underline{cuts} is also a negated event (which might not take place) while the \underline{rout} was further intensified (with the cue word: extending) and \underline{surplus} is ``eased'' (with the cue word: erase). Therefore, it is incomplete to just represent events through event extraction without accurate event properties classification. 

In this work, we look into the training of event extraction and event classification on crude oil-related events on \textit{CrudeOilNews} corpus. The contributions of this work is summarized as follows:
\begin{enumerate}
    \item Proposed an end-to-end Event Extraction solution placing equal emphasis on accurate event properties classification apart from event extraction;
    \item Utilize Transfer Learning to improve final model performance through improving embeddings or model representations to overcome issues of labeled data scarcity and class imbalance.
\end{enumerate}

%% Backup examples:
%%(2) The market has also been supported by indications that the OPEC + alliance is closing in on a deal to delay a planned easing of output cuts.
%%(3) fears that rising U.S. output would dampen OPEC 's efforts to rid the market of excess supplies prevented prices from rising much further

The rest of the article is organized as follows: In Section \ref{sec:related_work}, we lay out related work according to each sub-task; in Section \ref{sec:framework}, we present our proposed framework for a complete event extraction for \textit{CrudeOilNews} corpus. We dive deeper into the respective sub-tasks in subsequent sections: Section \ref{sec:event_extraction} covers event extraction  while Section \ref{sec:event_properties} focuses on event properties classification. Lastly, we present the conclusion in Section \ref{sec:conclusion}. 

\section{Related Work} \label{sec:related_work}
This section is structured such that related work will be discussed based on the specific area of interest, giving a targeted analysis of each area or sub-task within the overall event extraction task. 
%For simplicity, only solutions using Deep Learning are considered here. 
%Overall transfer learning in NLP is explored in \cite{mou2016transferable}

\subsection{Event Extraction} \label{subsec:EE_transferLearning}
The question of how to improve the extraction accuracy from a somewhat limited set of labeled gold data has become an important one. Recently many have started exploring transfer learning to improve event extraction through various types of Transfer Learning described below:

%(target task) by leveraging knowledge from training a related source task. 

%\paragraph{Unsupervised Learning} s. Peng et al. (2016) first attempted to extract event triggers with minimal supervision using similarity-based heuristics. 

\paragraph{Multi-task Learning (MTL)}
Multi-task Learning is also known as \textit{joint learning} or \textit{joint training} in most event extraction literature. Here we list the past work according to the different combinations of sub-tasks:
\begin{enumerate}
    \item Partial Multi-task learning: jointly training ED + APR. This approach is very common in event extraction literature and is found in \cite{lee2021effective, liu-etal-2018-jointly, li2013joint, nguyen2016joint, Sha_Qian_Chang_Sui_2018}. Although the approach of jointly training ED and APR is the same, all of them used different deep learning architecture. \cite{lee2021effective} used Graphical Convolution Network (GCN) + Pruned Dependency Parse Tree,  \cite{liu-etal-2018-jointly} used GCN with Attention Mechanism, \cite{nguyen2016joint} uses Recurrent Neural Network (RNN),
    \cite{Sha_Qian_Chang_Sui_2018} uses Dependency-Bridge RNN and Tensor-Based Argument Interaction. 
    %In \cite{lee2021effective} the multi-task learning approach was used where event trigger extraction and event argument extraction were trained concurrently.
    \item Full Multi-task learning: Joint modeling of all three sub-tasks : EMD, ED and ARP extraction. This approach was in reported in \cite{li-etal-2014-constructing, yang-mitchell-2016-joint, judea-strube-2016-incremental, nguyen2019one, zhang2019extracting}. The authors in \cite{yang-mitchell-2016-joint} consider structural dependencies among sub-tasks by adopting a two-stage reranking procedure, first selecting the k-best output of event triggers and entity mentions, then performing joint inference via reranking. \cite{nguyen2019one} build a multi-task model that exploits mutual benefits among the three tasks by sharing common encoding layers given an input sentence. In this setting, output structures of entity mentions, event triggers, and argument semantic roles are decoded separately. \cite{zhang2019extracting}, on the other hand, used a neural transition-based framework to predict complex joint structures incrementally in a state-transition process.
    \item Hierarchical Multi-task learning: training sub-tasks according to a hierarhical fashion. The idea is to utilize a set of low-level tasks learned at the bottom layers of the model to create a set of shared semantic representations that will progressively have a more complex representation from the more complex tasks at the higher-level. The authors in \cite{sanh2019hierarchical} showed that these low and higher level tasks benefit trained in a hierarchical fashion benefit each other. The authors trained Named Entity Recognition (NER), EMD, Entity Coreference Resolution, and Relation Extraction via a hierarchical fashion. Similarly, authors in \cite{wadden-etal-2019-entity} also aims to train the same set of sub-tasks (but treating entity co-reference as an auxiliary task) using span representation from BERT (and Graph propagation).
\end{enumerate}
Note: Multi-task solutions involving auxiliary tasks other than event extraction, for example, the combination of \textit{ED + Entity Relation Extraction} is excluded here. 
%In \cite{el2021mttlade}, on the other hand, uses multi-task transfer learning for Entity Mention + Entity Relation in adverse drug event extraction
%% Check out https://github.com/BaptisteBlouin/EventExtractionPapers for complete list of EE papers.

\paragraph{Sequential Transfer Learning (STL)}
According to \cite{ruder2019neural}, STL is a type of transfer learning. In this approach, We train a model on a task or a dataset, and then `transfer' the model to another task or dataset. This means that, as opposed to MTL, STL models are not optimized jointly, but each task is learned sequentially. \cite{chen2021} is an example of cross-domain STL; the authors used multiple source datasets to help achieve a wider coverage of events in the target dataset using adversarial network-based transfer learning. The authors capitalized on four other corpora with varying degrees of relevance to their target dataset (all within the BioMedical domain) to extract and transfer common features from the related source corpora effectively to boost the performance of event trigger detection in the target dataset.

%In \itep{ollagnier2020sequential}, the authors used cross-task STL where pre-trained language model like ELECTRA, DilstilBERT, OpenAI GPT2, and etc, were used for event detection and key sentence extraction. The authors have identified this as STL while in other work this is more commonly known as \textit{Fine-Tuning} from a pre-trained language model. 
%According to \cite{alyafeai2020survey}, \textit{Fine-Tuning} and \textit{Zero-shot} are two of the four categories under STL. 

\paragraph{Other forms of Transfer Learning}
In \cite{huang-etal-2018-zero}, the authors used zero-shot transfer learning to allow their event extraction model to generalize to new unseen event types (events without annotation). They model event extraction as a generic grounding problem and designed a transferable architecture of structural and compositional neural network, that leverages existing event schemas and human annotations for a small set of seen types, and transfers the knowledge from the existing types to the extraction of unseen types. In \cite{lyu2021zero}, on the other hand, the authors formulate \textit{zero-shot} event extraction as a set of Textual Entailment (TE) and / or Question Answering (QA) queries, exploiting pretrained (TE/QA) models for direct transfer (transfer learning) to do the new target task of event extraction. 

\subsection{Event Properties Classification}
Even though the ACE2005 dataset is annotated with not just event details but also properties such as \textit{Polarity, Tense, Genericity, and Modality}, previous work within the ACE2005 and TAC-KBP stream focused almost exclusively on event detection and event extraction and under-utilizing the annotation on event properties. Even in survey papers such as \cite{8918013} and \cite{hogenboom2016survey}, the focus is solely on event extraction, omitting event properties classification and other event-related task. Instead, event properties related tasks are established separately from event extraction through several shared tasks that are not necessarily event extraction related. These tasks come in various variations and different focuses, they are: 
\begin{enumerate}
    \item Event Realis classification \cite{mitamura2015event} in TAC KBP dataset. There are three types of Ralis values: {\fontfamily{qcr}\selectfont ACTUAL}, {\fontfamily{qcr}\selectfont GENERIC}, and {\fontfamily{qcr}\selectfont OTHER}. This is equivalent to our Modality property;
    \item CoNLL-2010 shared task: Hedge detection and scope resolution. The task is to detect hedges and their scope in natural language text. A detailed task description is found in \cite{farkas2010conll}. This is equivalent to our Modality property;
    %\footnote{Hedge detection mentioned here is synonymous with Uncertainty detection} 
    \item SEM 2012 Shared Task: Negation detection and scope resolution. The task is to detect negation and resolve its scope and focus. A detailed description of the task is found in \cite{morante2012sem}. This is equivalent to our Polarity property;
    \item Modal sense classification \cite{marasovic2016multilingual}. This is similar to Uncertainty hedge / Modality cue word detection; 
    \item Event Factuality Prediction (EFP) \cite{sauri2009factbank}. This is a combination of Negation and Speculation detection but instead of classification, EFP is a regression task to predict a score between [+3, -3]. 
\end{enumerate}

Even though none of the corpora above are related to Economic / Finance domain, their tasks are similar to our event property classification. Therefore they are potential resources for consideration for using them as resources as source datasets in cross-domain STL. 

%Even though corpora such as ACE2005, TAC KBP datasets and Richer Event Description (RED dataset) has a lot of information annotated but many work has chosen to focus solely on event detection and event extraction. This is evident in the event extraction survey paper by \cite{8918013} and \cite{hogenboom2016survey} that focused exclusively on event extraction methods while excluding any event Properties / properties classification.  

%In comparison to event extraction, event properties classification has received comparatively lesser focus and has fewer publications. One of the work -\cite{lu2021constrained} uses TAC KBP Realis classification as one of the task in their multi-task learning. Their setup of multi-task involves realis classication + event + entity coreference resolution.

\section{Proposed Framework} \label{sec:framework}
We present a framework for end-to-end event processing that includes both event extraction and as well as event properties classification. First, we build baseline models via supervised learning for each sub-task using only the \textit{CrudeOilNews} dataset. Then we explore various transfer learning techniques and leverage on available resources (source task or source dataset) to train target models with better performance. We demonstrate, through experiments, how much improvement (if any) these new models show compared to the baseline models.  

\begin{figure*}[h]
\centering
    \includegraphics[width=0.95\textwidth]{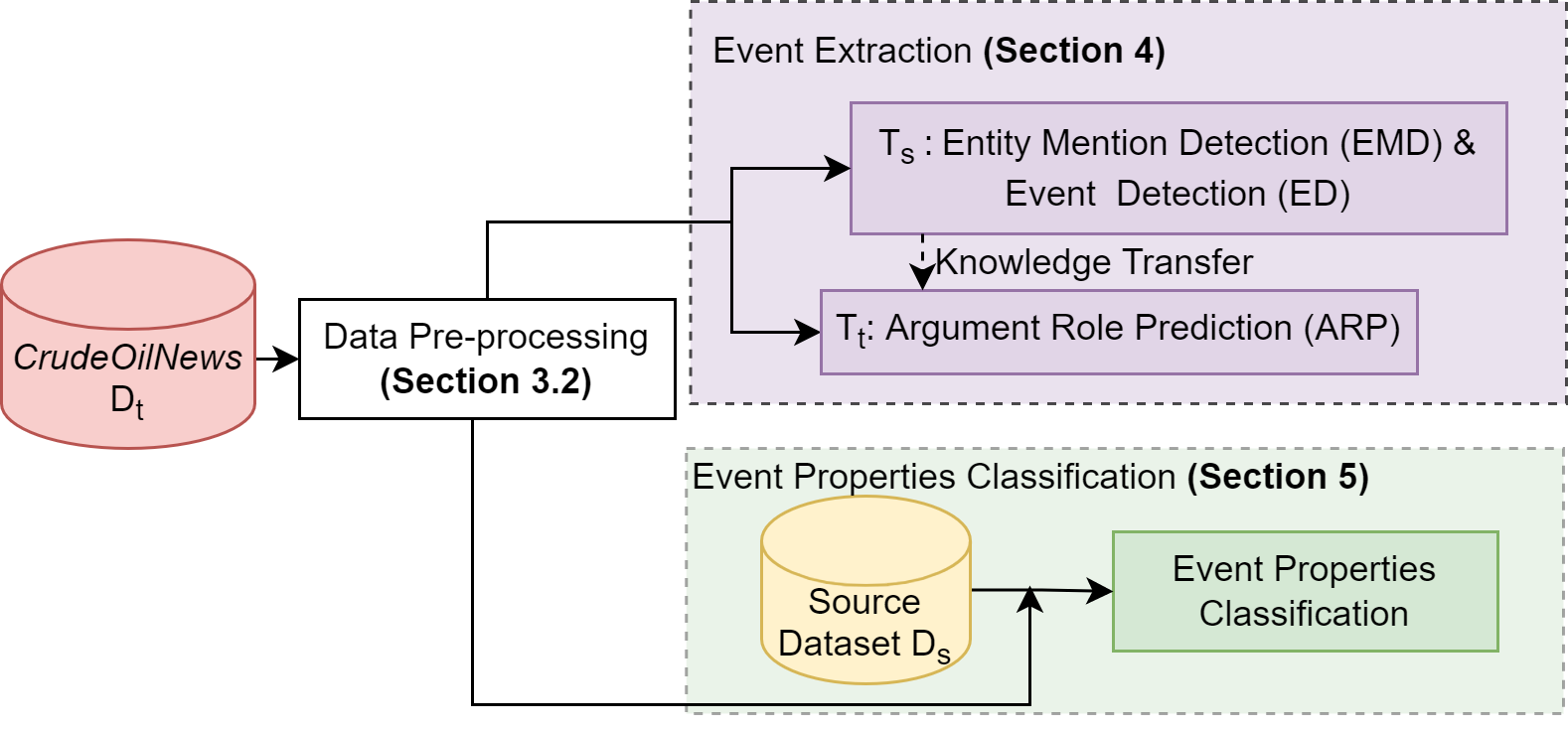}
    \caption{The framework is made up of different components, each corresponding to the sub-tasks in EE. Event extraction (EMD, ED and ARP) is covered in Section \ref{sec:event_extraction} while event properties classification (Polarity, Modality and Intensity) is covered in Section \ref{sec:event_properties}.}
    \label{fig:overall_framework}
\end{figure*}

By referring to Figure \ref{fig:overall_framework}, the proposed framework is described in detail in the following sections: 
\begin{enumerate}
    \item Domain Adaptive Pre-training on BERT to produce ComBERT for contextualized word embedding in Section \ref{sec:comBERT}
    \item Data pre-processing in Section \ref{sec:data_processing};
    \item Event Extraction (EE) and its sub-task: (1) Entity Mention (EMD) and Event Detection (ED) and (2) Argument Role Prediction (ARP) in Section \ref{sec:event_extraction}. Here, we explore the usage of cross-domain sequential transfer learning and inductive transfer learning.
    \item Event properties classification in Section \ref{sec:event_properties}. We explore cross-task sequential learning.
\end{enumerate}

\subsection{ComBERT: Domain Adaptive Pre-training} \label{sec:comBERT}
%The proposed solution adopts a shared representation obtained from the pre-trained language model learned through transformer architecture and ends up with task-specific fine-tuning. 
\begin{figure*}[h]
\centering
    \includegraphics[width=\textwidth]{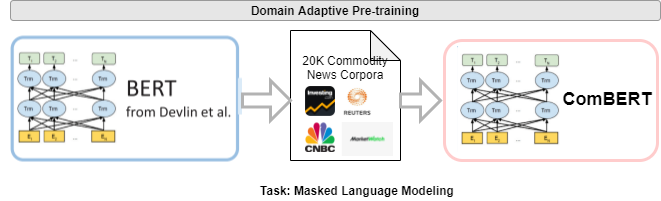}
    \caption{Domain Adaptive Pre-training: Using BERT as baseline, we have further pre-trained BERT on a commodity news corpus, adapting the model to the finance and economic news domain.}
    \label{fig:ComBERT}
\end{figure*}

As for tokens, the model takes an input sentence as a sequence of word tokens encoded through ComBERT, as introduced in \cite{lee2021effective}. As shown in Figure \ref{fig:ComBERT}, ComBERT is produced from BERT-based domain adaptive pre-training on a large collection of commodity news on the task of masked language modeling. It was trained on a collection of commodity news extracted from \href{investing.com}{www.investing.com}, the same source as the \textit{CrudeOilNews} corpus. The idea is to adapt BERT, which was trained on English Wikipedia and Brown corpus \cite{devlin-etal-2019-bert} to the finance and economics domains via Domain Adaptive Pre-training. In \cite{lee2021effective}, it is shown that ComBERT produced better event extraction results than BERT and other generic language models. This is consistent with language models produced through domain adaptive pre-training on in-domain corpus such as SciBERT \cite{beltagy2019scibert} and BioBERT \cite{lee2020biobert}. All event extraction tasks described later use ComBERT embeddings and models are trained via fine-tuning from ComBERT. 

\subsection{Data Preprocessing} \label{sec:data_processing}
The annotation files made public in \cite{lee2022crudeoilnews} were first converted from Brat Annotation standoff format (\textit{.ann} files) along with their corresponding news articles (\textit{.txt} files)  to \textit{json} format. Each sentence in the dataset was parsed using Stanford CoreNLP toolkit \cite{manning2014stanford}, including sentence splitting, tokenization, POS-tagging (lexical information), NER-tagging, and dependency parsing to generate dependency parse trees (syntactic information).
For all the sub-tasks, we adopt the ``multi-channel'' channel strategy where apart from using word embeddings from ComBERT, we include, as input, additional information on each token, ie. their POS tag, NER tag and dependency parse tree tags. 

\paragraph{Input} Let \textit{W} = $ w_1, w_2,.....w_n $ be a sentence of length n where $ w_i $ is the \textit{i}-th token:
\begin{enumerate}
    \item The word embedding vector of $w_i$: this is the feature representation from a word embedding of ComBERT. It is made up of WordPiece tokenization \cite{DBLP:journals/corr/WuSCLNMKCGMKSJL16} with [CLS] and [SEP]\footnote{[CLS], [SEP], [MASK] are special tokens of BERT. For experiments involving RoBERTa, Byte-Pair Encoding (BPE) tokenization and its special tokens are used.} are placed at the start and end of the sentence.
    \item The Part-of-Speech-tagging (POS-tagging) label embedding vector of $w_i$: This is generated by looking up the POS-tagging label embedding.
    \item Dependency tags: The same "Multi-channel" approach is used where the tokens' Universal Dependency dependency tags embedding are used. Examples of tags are \textit{nmod, nsubj, dobj, neg}, etc. This is based on the fact that entity mentions do not influence the Polarity and Modality of an event. On the other hand, modifier words and modal auxiliary such as \textbf{not}, which has the dependency tag \textit{neg}, are key to determining the Modality / Polarity / Intensity of an event. Dependency tags provide additional syntactic information, which is helpful in the classification task. 
\end{enumerate}

\subsection{Train-Test split}
Due to the limited size of \textit{CrudeOilNews} corpus, we run the experiments by using 5-fold cross-validation. Out of the 5-folds, one fold is for testing while the remaining four folds are for training (80\% for training and 20\% for testing). 

\subsection{Measurement for dataset with class imbalance}

\paragraph{\textbf{F1-Score}}
F1-score reported here is macro-average F1-score averaged across $k$ experiments. We compute the F1 score for each fold (iteration); then, we compute the average F1 score from these individual F1 scores.

\begin{equation}
 F1_{avg} = \frac{1}{k}\Sigma_{i=1}^k F1_i
    \label{eq:1}
\end{equation}

\paragraph{\textbf{MCC}}
In \cite{xie2013semantic}, apart from the familiar F1-measurement the authors used an additional evaluation metric known as the Matthew Correlation Coefficient (MCC) to avoid bias due to the skewness of data. It takes into account true and false positives and negatives and is generally regarded as a balanced measure, which can be used even if the classes are of very different sizes. MCC is a single summary value that incorporates all four cells of a $2 x 2$ confusion matrix\footnote{For more information on Matthews Correlation Coeffecient (MCC), visit}. %\href{https://scikit-learn.org/stable/modules/model_evaluation.html}{https://scikit-learn.org/stable/modules/model_evaluation.html}}: 

The equation for Binary Classification:
\begin{equation}
 MCC = \frac{TP \times TN - FP \times FN}{\sqrt{(TP +FP)(TP +FN )(TN +FP)(TN +FN )}}
    \label{eq:adfasdf}
\end{equation}

\noindent and for Multi-class Classificatioin:
\begin{equation}
 MCC = \frac{c \times s - \Sigma_k^K p_k \times t_k}{\sqrt{(s^2 - \Sigma _k^K p^2_k) \times (s^2 - \Sigma_k^K t^2_k)}}
    \label{eq:adfasdf}
\end{equation}
with the following intermediate variables:
\begin{itemize}
\setlength\itemsep{-0.3em}
    \item $t_k = \Sigma^K_i C_{ik}$ is the number of times class $k$ truly occurred,
    \item $p_k = \Sigma^K_i C_{ki}$ is the number of times class $k$ was predicted,
    \item $c = \Sigma^K_k C_{kk}$ is the total number of samples correctly predicted,
    \item $s = \Sigma^K_i \Sigma^K_j C_{ij}$ is the total number of samples,
    \item $TP$ is True Positive, $FP$ is False Positive, $TN$ is True Negative and $FN$ is False Negative.
\end{itemize}

\section{Event Extraction} \label{sec:event_extraction}
In this section, we first build baseline models for all event extraction subtasks in Section \ref{subsec:ee_baseline}. Then we investigate the extent of how transfer learning improves the performance of Event Extraction models from our baseline models. First, we look at cross-domain STL in Section \ref{subsec:EE_STL}, and then Inductive Transfer Learning can be used among the sub-tasks in Section \ref{subsec:inductive}.

\subsection{Baseline Model} \label{subsec:ee_baseline}
First, we start with our baseline model, where we train each sub-task individually. The model architecture of each sub-task is described below:

\subsubsection{Model Architecture}

\paragraph{\textbf{Entity Mention Detection (EMD) \& Event Trigger Detection (ED)}}
We formalize both Entity Mention Detection Model (EMD) and Event Trigger Detection (ED) tasks as  multi-class token classification where a token is classified as being at the beginning, inside, or outside an entity mention/event trigger (BIO notation). Similar to the approach used in \cite{nguyen2016joint}, we employ BIO annotation schema to assign entity type labels to each token in the sentences. For the model architecture, we use Huggingface's  \texttt{BERTForTokenClassification} with ComBERT embedding to fine-tune on this task. Both EMD and ED are trained separately on minimizing the cross-entropy loss function. 

\textbf{Pipeline versus Golden Entity Mention:}
A large portion of prior work on Event Extraction has taken a simplified approach that only focused on ED and ARP, ignoring the training of EMD \cite{ li2013joint, chen2015event, nguyen2016joint, lee2021effective, liu-etal-2018-jointly}. Experiments for ED and ERP are ran based on golden annotation for entity mentions as input. Here in this work, we aim to have a more realistic approach where we start with the EMD sub-task and train other sub-tasks based on the entity mentions predicted by the EMD model. We explore various task setups, i.e. various combinations of single task, multi-task joint training, and sequential transfer learning to minimize the issue of error propagation inherent in the pipeline approach. Details of the various task setups are covered in Section \ref{subsec:inductive}. 

\paragraph{\textbf{Argument Role Prediction (ARP)}}
We use the solution proposed by \cite{lee2021effective}, where each sentence's syntactic information (the dependency parse tree) is used to assist with the classification of event argument roles even though they are identical in terms of entity type, which also includes the abundant numerical values that play different event roles. Specifically, the proposed solution uses the pruned dependency parse tree along with Graphical Convolution Network (GCN) as the architecture. 

\begin{figure}[h]
\centering
    \includegraphics[width=0.55\textwidth]{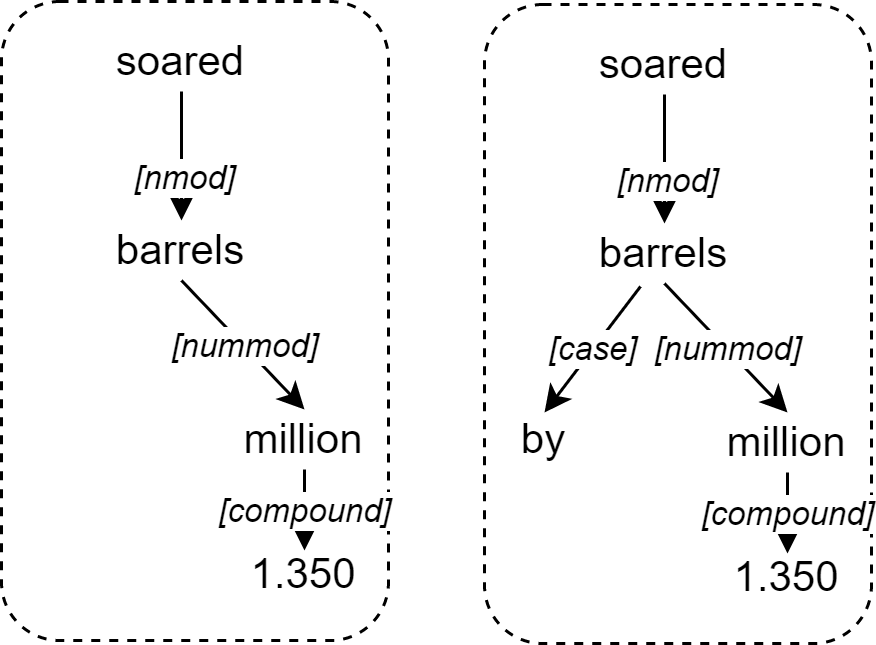}
    \caption{Left: pruned dependency parse with the shortest path between candidate trigger-entity. Right: pruned dependency parse tree with $k=1$ distance with additional contextual information for the classification of argument roles.}
    \label{fig:parse_trees}
\end{figure}

The architecture for the task of Argument Role Prediction (ARP) is similar to that of \cite{lee2021effective}, which uses Graph Convolutional Network (GCN) with pruned contextual parse tree. Here we explain what pruned contextual parse tree is and how to encode this pruned tree to be used as input in the ARP training. As part of data pre-processing, we use Stanford CoreNLP to generate a dependency parse tree for each sentence. We pruned the dependency tree to obtain a sub-tree between each candidate trigger with each candidate entity within a sentence. To further provide more contextual information, the dependency parse tree is pruned based on $k$ distance away. Figure \ref{fig:parse_trees} shows the comparison between a pruned tree of the shortest path of the candidate trigger-entity pair and a pruned tree of the same pair but pruned to include additional words $k$ distance away. This task is trained on minimizing the cross-entropy loss. 

This task is trained to minimize cross-entropy loss. For technical details of the implementation of Graph Convolutional Network (GCN) over pruned contextual parse tree, see \cite{lee2021effective}. 

\subsubsection{Experiments}

The result of the baseline solution is shown in Table \ref{table:result_sequential} (see first row), with results for individual sub-tasks under their respective header.

\subsection{Cross-domain Sequential Transfer Learning} \label{subsec:EE_STL}
%% Sequential transfer learning from other source domain

Inspired by the works of \cite{meftah2018neural} of using cross-domain STL in transferring model trained on POS-tagging task from Newswire domain (source domain) to Twitter text (target domain), here we investigate if we can utilize available source datasets on event extraction to improve performance of the same task in our target dataset further, \textit{CrudeOilNews}. There are two event extraction datasets annotated according to the ACE/ERE standards:  (1) benchmark event extraction dataset ACE2005 in the generic domain, and (2) SENTiVENT \cite{jacobs2021sentivent} for company-specific events in the finance and economics domain. However, unlike \cite{chen2021}, the source datasets that they used are basically from the same domain, i.e., BioMedical Domain. In our case, we do not have any other event extraction corpus from the same domain as \textit{CrudeOilNews}. The only two candidate corpus identified above are rather different from our target dataset; analysis for each one is listed down below: 

\paragraph{\textbf{ACE2005}} is a general domain corpus; out of its 33 sub-event types, almost none overlap with the events defined in \textit{CrudeOilNews} corpus. Even though 2 of the events \textit{Conflict - Attack, Conflict - Demonstrate}) may seem the same as \textit{Civil-unrest}, however upon closer scrutiny, the types of conflict here are rather different: ACE2005 ones are at a personal level, such as a person attacking another person, while in \textit{CrudeOilNews} the conflicts are geo-political, such as social unrest, large-scale demonstration.

\paragraph{\textbf{SENTiVENT}} is a corpus made up of business news and its event types, mainly company-related or company-level events. Among the event types, there is a `placeholder' event type called `Macroeconomic', a broad category that captures all non-company specific events such as market trends, market-share, competition, regulation issues, etc. This `Macroeconomic' event type is the only event type that overlaps with \textit{CrudeOilNews} corpus, unfortunately, while they lump non-company events into one category, \textit{CrudeOilNews} corpus focuses on Macro-economic and Geo-political events in a finer-detail. Furthermore, \textit{SENTiVENT} corpus is annotated with discontinuous, multiword triggers, e.g., ``upgraded ... to buy'', ``cut back ... expenses'', ``EPS decline''). This is distinctly different from the way triggers are annotated in \textit{ACE2005} and \textit{CrudeOilNews} where triggers are single-word or continuous multiwords. The baseline model developed for event detection in \textit{CrudeOilNews} cannot be readily applied to \textit{SENTiVENT} without any modification. 

Based on the fact that there is minimal overlap of event types between the candidate corpus above and our target dataset, we conclude that we are not able to utilize these two candidate corpora for cross-domain STL. This observation is supported by the results in \cite{chen2021}, where two out of four of the source dataset has a very low proportion of trigger overlap that produced worse performance in the target dataset. This is the result of Negative Transfer.

\subsection{Inductive Transfer Learning} \label{subsec:inductive}
%% Inductive Transfer Learning within Target Dataset
As shown in the section above, we are not able to utilize any available source datasets to boost task model performance via cross-domain STL. Instead, here we explore Inductive Transfer Learning, where we experimented with multi-task learning (MTL) and sequential transfer learning (STL) and the combination of the two among event extraction sub-tasks within our own dataset to obtain the best possible model performance. In the experiment section, we investigate \textbf{Single Task (baseline) vs Multi-task Learning (MTL) vs Sequential transfer Learning (STL) vs a combination of MTL and STL}.
%Sequential Transfer Learning is also known as Model Transfer in \cite{wang2015transfer}. 

\begin{figure*}[h]
\centering
    \includegraphics[width=1\textwidth]{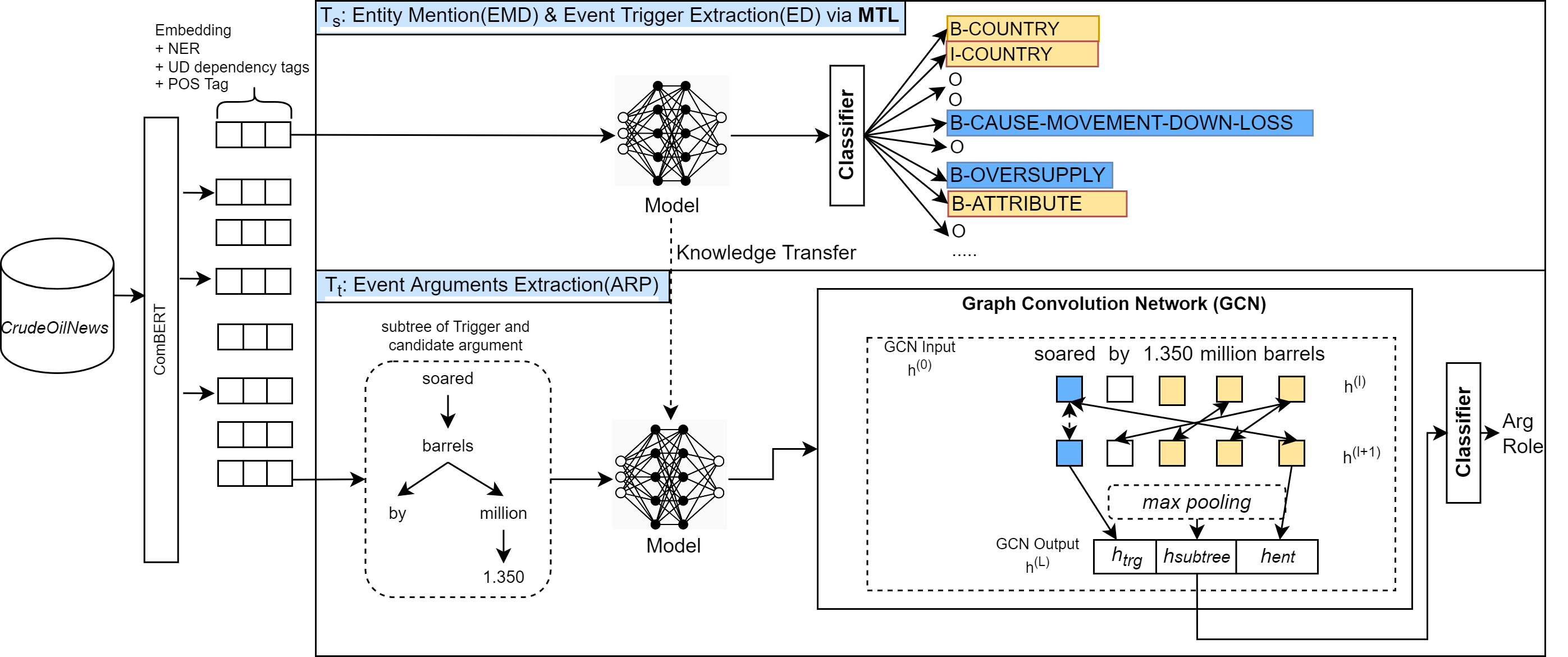}
    \caption{The combination of MTL and STL: First, the model is jointly trained on Entity Mention (EMD) and Trigger Extraction (ED) tasks via MTL and then the model is transferred sequentially to train on the task of Entity Arguments Extraction (ARP).}
    \label{fig:sequential_transfer_learning}
\end{figure*}

As observed by authors in \cite{sanh2019hierarchical},
%and \cite{lu2021constrained}, 
hierarchical multi-task transfer learning in a neural network typically allows the different tasks involved to benefit from each other via sharing the learned representations. 
STL consists of two stages: a pre-training phase in which general representations are learned on a \textit{source} task or domain, followed by an adaptation phase during which the learned knowledge is applied to a target task or domain. Figure \ref{fig:sequential_transfer_learning} shows the `model transfer' from the \textit{source task} - EMD + ED (top box) to the \textit{target task} - ARP (top box). 

\subsection{Experiments and Analysis}
We carried out five types of experiments with different combinations of task setups to determine the best transfer learning configuration with maximum benefits in terms of sharing learned representations of source tasks and target tasks. These five different task setups are:
\begin{enumerate}
    \item Single Task Learning (Baseline): this is also known as the pipeline approach where the sub-tasks are trained independently one after another, each model is trained from scratch (no transfer learning): EMD, ED, ARP.
    \item Full Multi-task training: For the experiment, we use the approach in \cite{zhang2019extracting}, where all sub-tasks are trained jointly using neural transition-based framework, predicting joint output structure as a single task.
    \item Full Sequential Training: in this approach, we start training for the EMD task and upon completion, transfer the model to train on ED and lastly on ARP:  EMD $\rightarrow$ ED $\rightarrow$ ARP. 
    \item Combination of MTL and STL: 
        \begin{enumerate}
            \item Combination \#1: EMD $\rightarrow$ ED + ARP. the approach of jointly training ED + ARP is very common in event extraction literature. This setup is used in \cite{lee2021effective}, and many others see Section \ref{subsec:EE_transferLearning}. The difference between \cite{lee2021effective} and this work is that in \cite{lee2021effective}, they jointly trained ED + ARP together using golden entity mentions, while in this work we use the more realistic setting where the input to the ED + ARP task is based on entities predicted from the earlier EMD + ED model (EMD $\rightarrow$ ED + ARP). 
            \begin{equation}
                 joint\_loss = loss\_EMD + \beta (loss\_ED)
            \end{equation}. In the experiment, we use $\beta$=2 to give a higher weightage to the loss of the ED task. 
            \item Combination \#2: EMD + ED $\rightarrow$ ARP as shown in Figure \ref{fig:sequential_transfer_learning}. The resulting model from training EMD + ED via MTL (in the upper box of Figure \ref{fig:sequential_transfer_learning}) is then transferred to train for ARP (lower box in the figure). EMD + ED acts as the source task $T_s$ in the context of cross-task STL to benefit the ARP task, the target task $T_t$. The joint loss for EMD + ED is as follows: 
            \begin{equation}
             joint\_loss = loss\_EMD + loss\_ED
                \end{equation}
        \end{enumerate}
\end{enumerate}

The results of these experiments are shown in Table \ref{table:result_sequential}. The detailed breakdown by entity mention type and event type are reported in Table \ref{table:entity_trigger_results} while results breakdown by event argument roles are reported in Table \ref{table:ARP_results}, both in \ref{app:additional_results}.

 \begin{table*}[h!] 
    \centering \small
    \caption{Experiments with various different task-setups for Event Extraction to investigate \textbf{Single Task Learning vs Multi-task Learning vs Sequential transfer Learning}. All setups use the list of entities extracted from the EMD task and not based on Golden annotation.}
    \begin{tabular}{ |p{0.35\linewidth} | p{0.8cm} | p{0.8cm} | p{0.8cm}| p{0.8cm} | p{0.8cm} | p{0.8cm} | p{0.8cm} | p{0.8cm} | p{0.8cm}| p{0.8cm} |}  \hline
     \multirow{2}{*}{\textbf{Task setup}} & \multicolumn{3}{c|}{\textbf{EMD}} & \multicolumn{3}{c|}{\textbf{ED}} & \multicolumn{2}{c|}{\textbf{ARP}}\\ \cline{2-9}
     & P & R & F1  & P & R & F1 & F1 & MCC  \\ \hline
     1. Individual sub-task training (Baseline) & 0.903 &  0.912 & 0.907 & 0.915 & 0.899 & 0.907 & 0.802 & 0.694\\ \hline 
     \multicolumn{9}{c}{\textbf{ }} \\ \hline
     
     \multirow{2}{*}{\textbf{Task setup}} & \multicolumn{8}{c|}{\textbf{EMD + ED + ARP}} \\ \cline{2-9}
     & P & R & F1  & P & R & F1 & F1 & MCC  \\ \hline
     2. Full Multi-task Training \cite{zhang2019extracting} & 0.879 & 0.891 & 0.880 & 0.901 & 0.905 & 0.904 & 0.854 & 0.710 \\ \hline 
     \multicolumn{9}{c}{\textbf{ }} \\ \hline
     
     \multirow{2}{*}{\textbf{Task setup}} & \multicolumn{8}{c|}{\textbf{EMD $\rightarrow$ ED $\rightarrow$ ARP}} \\ \cline{2-9}
     & P & R & F1  & P & R & F1 & F1 & MCC  \\ \hline
     3. Full Sequential Task Training & 0.879 & 0.891 & 0.880 & 0.901 & 0.905 & 0.904 & 0.854 & 0.710 \\ \hline 
     \multicolumn{9}{c}{\textbf{ }} \\ \hline
     
     \multirow{2}{*}{\textbf{Task setup}} & \multicolumn{8}{c|}{\textbf{EMD $\rightarrow$ (ED + ARP) }} \\ \cline{2-9}
     & P & R & F1  & P & R & F1 & F1 & MCC  \\ \hline
     4. Combination \#1: EMD $\rightarrow$ ED+ARP \cite{lee2021effective}\protect\footnotemark & 0.903 & 0.912 & 0.907 & 0.905 & 0.890 & 0.902 & 0.833  & 0.723\\ \hline 
     \multicolumn{9}{c}{\textbf{ }} \\ \hline
     
     \multirow{2}{*}{\textbf{Task setup}} & \multicolumn{8}{c|}{\textbf{(EMD + ED) $\rightarrow$ ARP}}  \\ \cline{2-9}
     & P & R & F1 & P & R & F1 & F1 & MCC  \\ \hline
     \textbf{5. Combination \#2: EMD + ED $\rightarrow$ ARP} & \textbf{0.926} & \textbf{0.937} & \textbf{0.931} & xx & xx & xx & \textbf{0.888} & \textbf{0.797} \\ \hline   %%EMD+ED iteration #31, ARP iteration #16
      \end{tabular} 
    \label{table:result_sequential}
\end{table*}
\footnotetext[7]{The results presented here is not a like-for-like comparison to the results presented in \cite{lee2021effective}. This is because in the paper golden entity mention was used as input to the joint-training of ED+ARP, while in this work, the output for EMD is used as input to joint-training of ED+ARP.}

\subsubsection{Analysis}
As expected, the worst-performing setup is the individual tasks-pipeline approach, where it not only suffers from error propagation, but each model is trained from scratch for each sub-task (without any interaction between them). Both full MTL and full STL achieved slightly better results. Between these two, the full multi-task training took a few more iterations and took longer to train because the approach is more complex.

%We use STL where the resulting model from EMD + ED task is used in the subsequent training of ARP. The reason behind doing this is to transfer knowledge from the source tasks (EMD + ED) to benefit the current target task (ARP). 

The best performing models are those that utilize a combination of MTL and STL task setups. Jointly training ED and ARP together (part of the combination \#1) is a common approach in the event extraction literature \cite{liu-etal-2018-jointly, lee2021effective}. However, in our case, we find that jointly training EMD + ED (combination \#2) brings better performance. This is because \textit{CrudeOilNews} does not exhibit strong interdependence between event type and argument roles as it is in ACE2005. This can be explained using example sentences found in the respective datasets in Table \ref{table:analysis_dataset}.

The best performing task setup is \textit{Combination \#2: EMD + ED $\rightarrow$ ARP}. The training of EMD and ED can be done jointly via MTL without much impact on both task. This is because we noticed that entity mentions and trigger words, by definition, are mutually exclusive, e.g. an entity such as \textit{crude oil} is never an event trigger, vice versa an event trigger such as \textit{glut}, though a noun, is never an entity mention. Treating EMD+ED as the source task is useful for the target task. This is related to the fact that the lower embedding and semantic information learned from joint training EMD + ED has a good level of knowledge about entities and triggers, the resulting model has a presentation that is useful for the ARP target task. 

 \begin{table}[h!] 
    \centering \small
    \caption{Analysis of events in \textit{ACE2005} and in \textit{CrudeOilNews} respectively. Both datasets exhibit different level of interdependence between event trigger words and its event arguments. The difference in the level of interdependence influenced the selection of the best MTL and STL combination for each of the dataset.}
    \begin{tabular}{ |p{0.40\textwidth} |p{0.60\textwidth} |}  \hline
    Dataset & Analysis \\ \hline
    \textbf{ACE2005:}
    
    (1) \textit{In Baghdad, a cameraman \underline{died} when an American tank \underline{fired} on Palestine Hotel}
    
    (2) \textit{He has \underline{fired} his air defence chief.} & The first occurrence of “fired” is an event trigger of type {\fontfamily{qcr}\selectfont ATTACK} while the second ``fired” takes {\fontfamily{qcr}\selectfont END-POSITION} as its event type. The authors in \cite{li2013joint} argues that event arguments play a key role in helping classifying the correct event. For example, the presence of ``tank'' plays the role of {\fontfamily{qcr}\selectfont WEAPON} helps determine the right event type. Likewise, in the second sentence, ``defence chief'' plays the role of {\fontfamily{qcr}\selectfont POSITION} can help the model classify the second ``fired'' as {\fontfamily{qcr}\selectfont END-POSITION}. Hence jointly training both ED and ARP helps improve the accuracy of both ED and APR.\\
    & \\ \hline
    \textbf{CrudeOilNews:} \textit{U.S. crude stockpiles \underline{soared} by 1.350 million barrels in December from a mere 200 million barrels to 438.9 million barrels, due to this \underline{oversupply} crude oil prices \underline{plunged} more than 50\% on Tuesday.} & Events are more straight forward, ie. trigger words are tied to just one type of event, therefore there is no need to utilize arguments to help differentiate the event type.  \\ 
    & \\ \hline
      \end{tabular} 
    \label{table:analysis_dataset}
\end{table}

\section{Event Properties Classification} \label{sec:event_properties}
\subsection{Baseline Model} \label{subsec:text_span}
%\subsubsection{Input Text Span} \label{subsec:text_span}
One of the challenges we face when training a model for event properties classification is to classify each event in sentences that contain multiple events accurately. According to \cite{lee2022crudeoilnews}, on average, the sentences in the \textit{CrudeOilNews} corpus has about 2 to 3 events. Accurate classification of event properties requires identifying cue words at the event scope level and not at the sentence level, i.e., classify the entire sentence. In comparison, both Negation and Uncertainty detection tasks in SEM2012 and CoNLL2011 are done at the sentence level. Therefore to accurately classify event properties, rather than using the whole sentence that may consist of several events, we have to narrow down the scope to use only the event scope, which is just a portion of the sentences. 

\subsubsection{Model Architecture}
To obtain this `scope', we experimented with the following inputs: 
\begin{enumerate}
    \item use the fixed window of words around event trigger word(s) ($x_{i-r}...x_{i-1} \oplus x_i \oplus x_{i+1}...x_{i+r}$) where $\oplus$ is the concatenation operation, $r$ represents the length from trigger word $x_i$. The sequential word representation is fed into an MLP to generate a vector and then through a \textit{softmax} activation function. 
    \item Use dependency parse sub-tree of event trigger in \cite{lee2021effective}  %but with modification
    \item Use \texttt{SelfAttentiveSpanExtractor} \cite{gardner2018allennlp} (part of the AllenNLP library) to weightedly combine the representations of multiple tokens and create a single vector for the original event span. The span vectors are fed into a two-layer feed-forward network with a \textit{softmax} activation function. 
\end{enumerate}

\subsection{Experiments} \label{subsec:prop_baseline}
\paragraph{\textbf{Train-Test Split}}
The main challenge in Event Property classification with the \textit{CrudeOilNews} corpus is class imbalance. To address this, we modified the k-fold cross-validation from random sampling to stratified sampling. This is done to ensure that both the training and testing set in each cross-validation maintain the same class distribution (label ratio) of the original dataset as shown in Figure \ref{fig:class_imbalance}.

\paragraph{\textbf{Results}}
All three sub-tasks (Polarity, Modality, and Intensity classification) are standalone and independent tasks where the outcome of one does not influence the outcome of others. Therefore all three classification models were trained independently of each other. Event Modality and Polarity classification is a binary classification task, the labels for Modality are: {\fontfamily{qcr}\selectfont ASSERTED} and {\fontfamily{qcr}\selectfont OTHER}, and for Polarity are: {\fontfamily{qcr}\selectfont POSITIVE} and {\fontfamily{qcr}\selectfont NEGATIVE}. Both classification tasks are trained on Binary Cross-entropy Loss. As for Event Intensity, it is a multi-class classification task, the labels are: {\fontfamily{qcr}\selectfont NEUTRAL}, {\fontfamily{qcr}\selectfont EASED}, and {\fontfamily{qcr}\selectfont INTENSIFIED}. It is trained on multi-class Cross-entropy Loss.
We run experiments to determine the most suitable text span for the classification of Event Property Classification by the text processing methods listed in Table \ref{table:baseline}.
 \begin{table}[h!] 
    \centering \small
    \caption{Experiment results of different Input Text Span.}
    \begin{tabular}{ |p{0.45\linewidth} | c| c | c |c|c |c|}  \hline
    \multirow{2}{*}{Text Span Generation Methods} & \multicolumn{2}{c|}{Polarity} & \multicolumn{2}{c|}{Modality} & \multicolumn{2}{c|}{Intensity} \\ \cline{2-7}
    & F1 & MCC & F1 & MCC & F1 & MCC \\ \hline
    $3$-grams fixed window centered around event trigger & 0.685 & 0.285 & 0.699 & 0.305 & 0.701 & 0.320 \\ \hline
    Dependency parse tree \cite{lee2021effective} & 0.759 & 0.305 & 0.723 & 0.298 & 0.699 & 0.298 \\ \hline
    SelfAttentiveSpanExtractor \cite{jiang2021he} & \textbf{0.892} & \small{\textbf{0.385}} & \small{\textbf{0.851}} & \textbf{0.462} & \textbf{0.751} & \textbf{0.665} \\ \hline
      \end{tabular} 
    \label{table:baseline}
\end{table}

\begin{figure}[ht!]
\centering
    \includegraphics[width=0.65\textwidth]{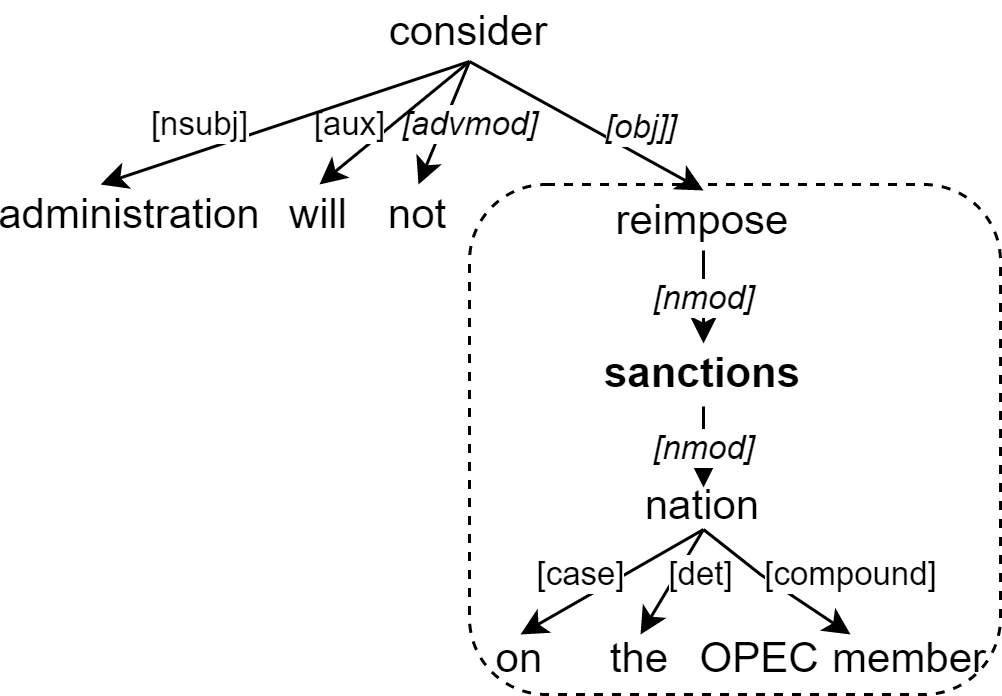}
    \caption{An example of a pruned dependency parse tree (enclosed in dotted lines) that did not generate a good text input for modality and polarity classification. Based on the event trigger word - \textbf{sanctions}, this pruned dependency parse tree does not contain the modality and polarity cue words: \textit{will} and \textit{not}.}
    \label{fig:negative_example}
\end{figure}

\paragraph{\textbf{Analysis}} 
Based on the results in Table \ref{table:baseline}, we conclude that the best text span is the ones generated by \texttt{SelfAttentiveSpanExtractor}, then followed by using a dependency parse tree. Dependency parse tree utilizes the syntactic structures of the input sentence and work well for identifying modifiers and negations such as WILL and NOT that are linked to the event trigger's sub-parse tree. However, it does not work for cases where event trigger is not a verb that forms its sub-tree. We illustrate this with an example sentence: ``The Trump administration will not consider reimposing sanctions on the OPEC member nation.'' and a portion of the dependency tree in Figure \ref{fig:negative_example}.

The worst performing text span is using neighboring words centering around event trigger because it does not capture words that are far away from the trigger (i.e.: words are located outside of the $3$-grams window). Based on the example above, the word cue word `will' is not extracted as part of the text span for \textbf{sanctions} because it is located outside of the fixed window centered around the event trigger.  In the section that follows, cross-domain sequential transfer learning is done using \texttt{selfAttentiveSpanExtractor} approach, the input text span that produces the best event property classification results.

The F1-score for both Polarity and Modality classification is high, while their respective MCC scores are much lower. This situation is caused by class imbalance in these two classifications, as shown in Figure \ref{fig:class_imbalance}. Error analysis on these two tasks showed the errors are primarily False Positives. The models tend to classify everything to the majority class ({\fontfamily{qcr}\selectfont POSITIVE} for Polarity classification and {\fontfamily{qcr}\selectfont ASSERTED} for Modality classification), which result in low precision. Given that MCC score takes into account all four values in the confusion matrix, a low MCC score shows that the models are not good at classifying the minority classes. 

The most challenging among the three tasks is Event Intensity classification. Some of the cue words for determining the event intensity ({\fontfamily{qcr}\selectfont NEUTRAL}, {\fontfamily{qcr}\selectfont EASED}, {\fontfamily{qcr}\selectfont INTENSIFIED}) are themselves trigger words. For example, \textit{\underline{Oversupply} could rise next year when Iraq starts to export more oil.} The correct interpretation should be: The event \underline{oversupply} might be further {\fontfamily{qcr}\selectfont INTENSIFIED} (cue word: \textit{rise}, but this word is also a trigger word for the event - {\fontfamily{qcr}\selectfont MOVEMENT-UP-GAIN}.  
\subsection{Cross-Domain Sequential Transfer Learning}
%%Sequential Transfer Learning from other source domain (previous header)

%usually performed in relation to processing negation: (1) negation cue detection, in order to find the words that express negation; (2) Scope identification, in order to find the words that express negation; (3) negated event recognition, to determine which events are affected by the negated cues and (4) focus detection, in order to find which parts of the scope that is most prominently negated. Dataset: \textbf{ConanDoyle-neg}

%The classification is done via a pipeline approach where an event trigger model is first trained. We use golden labels for event triggers for this task - Polarity / Modality classification. 

\begin{figure*}[ht!]
\centering
    \includegraphics[width=0.95\textwidth]{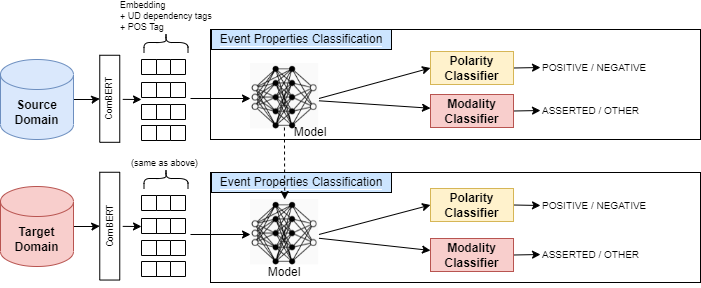}
    \caption{\textit{Sequential Transfer Learning (STL)}: The model is first trained on labeled dataset in the source domain (see list of corpora listed in Table \ref{table:transferLearning_corpora}) before being transferred to train on the Target Domain (\textit{CrudeOilNews}). STL is done for Polarity and Modality Classification. On the other hand, intensity classification is trained from scratch with just the \textit{CrudeOilNews} corpus due to the lack of resources.}
    \label{fig:transfer_learning}
\end{figure*}

Here we investigate the usage of other available corpora in all domains for the purpose of cross-domain STL. The idea is to use resources from different source domains to train a model on a task before fine-tuning the model to adapt to a new domain on the same task. We carry out this STL by first training the corpora from the \textit{source domain} on Event Polarity / Modality classification and then transfer the model to fine-tune on the same task on the \textit{target domain}, i.e., \textit{CrudeOilNews} corpus. Figure \ref{fig:transfer_learning} shows a graphical depiction of the idea of STL. In the top section, a labeled dataset in the source domain is used to train a model on event polarity or modality classification. The model is then trained using the dataset from the target domain and fine-tuned on the same task, as shown in the bottom section. 

Before going into the implementation details of cross-domain STL, it is important we align the task definitions as they may appear as different names: 
\begin{enumerate}
    \item \textbf{Event Polarity Classification}\footnote{Not to be confused with Sentiment Polarity(Positive / Negative sentiment)} can be aligned to \textbf{Negation Detection}, such as in \textit{SEM 2012 Shared Task: Negation Detection and Scope Resolution}. 
    \item \textbf{Event Modality Classification} can be aligned to \textbf{Hedge / Uncertainty / Speculation Detection} such as in \textit{CoNLL 2010 shared task: Hedge detection and scope resolution}. 
\end{enumerate}
In the subsections that follow, we lay out the available resources for both negation detection and uncertainty detection. Resources for Event Factuality Prediction (EFP) is excluded here because EFP combines both negation and speculation detection to determine the `factuality' of an event. 
%The EFP task is formulated as a regression task to predict a real score in the range of [-3, +3] to quantify the occurrence possibility of a given event mention. 
Instead, we look to corpora that have Negation and Uncertainty annotated individually to match how event polarity and modality are annotated in \textit{CrudeOilNews} corpus. 

\paragraph{\textbf{Corpora for Negation Detection}} 
\begin{enumerate}
    \item In SEM2012 Shared Task \cite{morante2012sem}, two corpora in the general text were released for negation scope and focus detection; they are the \textit{Conan Doyle stories} and the \textit{Penn TreeBank} corpus;
    \item In the survey paper \cite{jimenez2020corpora}, the authors listed out all English and Spanish corpora annotated with negation (negation cue and its respective scope). According to the list, majority of the available corpora are in the following domains: 
    \begin{enumerate}
        \item Bio-related text domain: \textit{BioInfer, Genia Event, BioScope, and DrugDDI};
        \item Consumer reviews: \textit{product review Corpus, SFU Review, Movie review};
        \item General Domain: \textit{Prop Bank} and \textit{SFU Opinion \& Comments (SOCC)}
    \end{enumerate}
\end{enumerate}

\paragraph{\textbf{Corpora for Uncertainty Detection}}
%Uncertainty detection is also known as \textit{hedge} or \textit{speculation} detection. 
\begin{enumerate}
    \item The CoNLL2010 share task \cite{farkas2010conll} is made up of a collection of corpora include Biology-related publications and general domain  factual text from Wikipedia;
    \item Financial domain, \cite{theil2018word} introduced the \textit{10-k financial disclosures corpus} for the task of classifying financial statements whether they are \textit{certain} or \textit{uncertain}.
    \item Consumer reviews : SFU Reviews corpus \cite{konstantinova2012review} contains both uncertainty and negation cue words and scope annotated. 
\end{enumerate}

\begin{table*}[h!]   
    \centering \small
    \caption{The list of open source corpora with Negation and Uncertainty Annotation.}
    \begin{tabular}{|p{0.35\linewidth}|p{0.20\linewidth} | p{0.10\linewidth} | p{0.10\linewidth}  |p{0.10\linewidth} |p{0.10\linewidth} |} \hline
    \multirow{2}{*}{\textbf{Dataset}} &
    \multirow{2}{*}{\textbf{Domain}} &
    \multicolumn{2}{c|}{\textbf{Negation}} & \multicolumn{2}{c|}{\textbf{Uncertainty}} \\
     \cline{3-6} 
     & & \textbf{Cue} & \textbf{Scope} & \textbf{Cue} & \textbf{Scope} \\ \hline
     1. ConanDoyle(neg) \cite{morante2012sem} & Fiction & \checkmark  & \checkmark &  &   \\ \hline
     2. SFU OCC(neg) \cite{kolhatkar2020sfu} & Opinion News \& Comments & \checkmark & \checkmark &  &  \\ \hline 
     3. 10kFinStatement(unc) \cite{theil2018word} & Corporate Financial Disclosure & & & \multicolumn{2}{l|}{\small{only class labels}}  \\  \hline
     4. Wikipedia(unc) \cite{farkas2010conll} & General &  &  & \checkmark &  \\ \hline
     5. Reviews(neg \& unc) \cite{konstantinova2012review} & Product Reviews & \checkmark & \checkmark & \checkmark & \checkmark \\ \hline
     6. SENTiVENT \cite{jacobs2021sentivent} & Economic news & \multicolumn{2}{c|}{only class labels} & \multicolumn{2}{c|}{only class labels} \\ \hline
    \end{tabular}
    \label{table:transferLearning_corpora}
\end{table*}

\subsubsection{Domain Similarity}
In transfer learning, it is observed that the more related the tasks, the easier it is for transfer or cross-utilize the knowledge \cite{sanh2019hierarchical}. The same holds true for data where the more related the source domain to the target domain, the easier it is for effective transfer learning \cite{meftah2021hidden, gururangan2020don}. Based on this, we excluded Bio-medical-related corpora from our experiments because the Bio-medical domain has its specific vocabulary deemed different from the Finance and Economics domain and is an ill-fit with ComBERT embeddings\footnote{see Section \ref{sec:comBERT} for more details about ComBERT}. Five corpora was selected and are listed in Table \ref{table:transferLearning_corpora}; the details of each of these corpora are found in \ref{app_dataset}.

\begin{figure}[ht!]
\centering
    \includegraphics[width=0.85\textwidth]{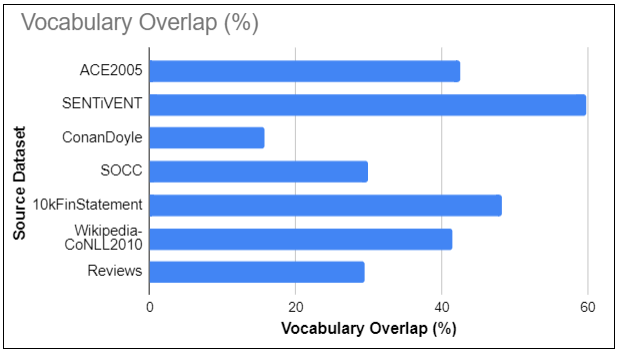}
    \caption{Vocabulary overlap (\%) between source datasets and \textit{CrudeOilNews} target dataset. Vocabularies of each domain are made up of the top 10K most frequent words (excluding stopwords) in each corpus.}
    \label{fig:vocabulary_overlap}
\end{figure}

We evaluate domain similarity between the source datasets and \textit{CrudeOilNews} by obtaining the percentage of vocabulary overlap of the two as shown in Figure \ref{fig:vocabulary_overlap}. On a continuum of the proximity between source datasets and target dataset, the source datasets can be ranked as \textit{SENTiVENT} $\rightarrow$ \textit{10kFinStatement} $\rightarrow$ \textit{Wikipedia-CONLL2010} $\rightarrow$ \textit{ACE2005} $\rightarrow$ \textit{SOCC} $\rightarrow$ \textit{Reviews} $\rightarrow$ \textit{ConanDoyle}.

\subsubsection{Task Modification}
The shared task of CoNLL2010 (for Uncertainty Detection) and SEM2010 (for Negation Detection) consist of two sub-tasks: (1) it involves first detecting the cue words at the sentence level and then (2) resolving the scope based on the cue words detected. Event property classification, on the other hand, is slightly different where the main aim is to detect the event first and then determine the event properties based on the event and its scope. 
%scope contains negation / uncertainty / intensity cue words. 
Due to the difference in the original source tasks and how the source datasets listed in Table \ref{table:transferLearning_corpora} are annotated, we modify the original tasks and adapted them to our task of event property classification by making the following modifications:
\begin{enumerate}
    \item Simplify the shared task to just one task. The original Negation / Uncertainy detection involves two sub-tasks : (1) cue word detection and (2) scope resolution. This is simplified into a single sequence classification task. 
    \item Align class labels: 
        \begin{enumerate}
            \item Event Polarity: For sentences that contain Negation cue words, we assign the label {\fontfamily{qcr}\selectfont NEGATIVE} for the whole sentence. For sentences without, we assign the label {\fontfamily{qcr}\selectfont POSITIVE} instead. 
            \item Event Modality: Similar to polarity classification, sentences with Uncertainty cue words or has the labeled `uncertain' are labeled with {\fontfamily{qcr}\selectfont OTHER}. For sentences without, we assign the label {\fontfamily{qcr}\selectfont ASSERTED}.
        \end{enumerate}
\end{enumerate}

\subsection{Experiments}
First, we train Event Polarity / Modality using the corpora listed in Table \ref{table:transferLearning_corpora}; these corpora are known as the ``source domain'' $D_s$. Then we transfer the model to train on the same task on the ``target corpus'' $D_t$, the \textit{CrudeOilNews} corpus. In our experiments, we use the best model trained on the source dataset for STL on \textit{CrudeOilNews}. In other words, we picked the model at the epoch with the highest performance on the source validation set. 

\paragraph{\textbf{Results}}
 \begin{table}[h!] 
    \centering \small
    \caption{The results of Polarity and Modality classification of \textit{CrudeOilNews} using various source datasets (identified in Table \ref{table:transferLearning_corpora}) as source domain $D_s$ in Sequential transfer learning.}
    \begin{tabular}{ |l | l| c | c |}  \hline
     \multicolumn{4}{|c|}{\textbf{Event Polarity Classification}} \\ \hline
    Source $D_s$ & Target $D_t$ & F1 & MCC \\ \hline
    1a. - & CrudeOilNews & \small{0.892} & \small{0.385} \\ \hline
    1b. ConanDoyle & CrudeOilNews & \small{0.699}($\downarrow$) & \small{0.305}($\downarrow$) \\ \hline
    1c. OCC & CrudeOilNews & \small{0.893}($\uparrow$) & \small{0.498}($\uparrow$) \\ \hline
    1d. Reviews & CrudeOilNews & \small{0.713}($\downarrow$) & \small{0.412}($\uparrow$) \\ \hline
    1e. \textbf{SENTiVENT} & CrudeOilNews & \textbf{0.895}($\uparrow$) & \textbf{0.678}($\uparrow$) \\ \hline \hline
    \multicolumn{4}{|c|}{\textbf{Event Modality Classification}} \\ \hline
    Source $D_s$ & Target $D_t$ & F1 & MCC \\ \hline
    2a. - & CrudeOilNews & \small{0.851} & \small{0.462} \\ \hline
    2b. 10kFinStatement & CrudeOilNews & \small{0.879}$\uparrow$ & \small{0.695}$\uparrow$ \\ \hline %% iteration 39
    2c. Wikipedia & CrudeOilNews & \small{0.841}($\downarrow$) & \small{0.501}($\uparrow$) \\ \hline
    2d. Reviews & CrudeOilNews & \small{0.723}($\downarrow$) & \small{0.395}($\downarrow$) \\ \hline
    2e. \textbf{SENTiVENT} & CrudeOilNews & \textbf{0.883}($\uparrow$) & \textbf{0.715}($\uparrow$) \\ \hline %\hline
    %\multicolumn{4}{|c|}{\textbf{Event Intensity Classification}} \\ \hline
    %Source & Target & F1 & MCC \\ \hline
    %3. - & CrudeOilNews &  0.711 & 0.765 \\ \hline  %%iteration 5
      \end{tabular} 
    \label{table:result_Properties_transferLearning}
\end{table}

Event Intensity classification is excluded from cross-domain STL because to the best of our knowledge, there is no available labeled dataset annotated for event intensity classification.

\paragraph{\textbf{Analysis}}

We analyze the results shown in Table \ref{table:result_Properties_transferLearning} by the list of event properties below:
\begin{enumerate}
    \item Event Polarity: there is some form of improvement when the model is trained on a source domain first before fine-tuning the model on the target domain on the same task. The best ``source domain'' corpus for Event Polarity is \textit{SENTiVENT}, while models trained on \textit{ConanDoyle} and \textit{Reviews} performed worst than the baseline model. The main reason is that these corpora are somewhat dissimilar to \textit{CrudeOilNews} corpus that resulted in Negative Transfer. Performance deterioration is especially apparent in \textit{ConanDoyle} because it is a corpus made up of dialogues or conversations and has negation cues mainly in a conversational form such as \textit{don't, doesn't, didn't, isn't, can't, wasn't} that are not found in the target corpus.
    
    \item Event Modality: Due to the similarity between the \textit{CrudeOilNews} and the two finance-related corpora : \textit{SENTiVENT} and \textit{10KFinStatement}, the resulting cross-domain STL models are able to provide a significant boost to model classification performance. Similar to the \textit{ConanDoyle} corpus, the \textit{Reviews} contains conversational-like sentences that have minimal overlap with \textit{CrudeOilNews} in terms of uncertainty cue words that resulted in model performance worse than the baseline. 
\end{enumerate}

It is worth highlighting that by comparing between F1-score and MCC-score, MCC-score has a more significant jump in improvement. As highlighted in Section \ref{subsec:prop_baseline}, results of baseline models show a lower MCC-score because of class imbalance; the models tend to classify everything to the majority class leading to a high number of False Positives. Upon training the model with source datasets that do not have a serious class imbalance issue, the final models have higher MCC-scores. Based on error analysis, it is shown that the final models have higher True Negatives (higher prediction on minority class) and thus lead to better MCC-score.

It can be concluded from the results above that the more similar the source domain is to the target domain, the easier we can use cross-domain STL to improve the final classifier's performance. It is also observed in \cite{ruder2019neural} that the more distant two domains are, the harder it is to adapt from one to the other. Apart from improving the models' F1 score, it is also observed that the models' MCC improves as well. An improved MCC score means classification performance improves across all classes, including the minority class, therefore addressing the issue of class imbalance.

%%%  Todo: compare before and after transfer learning
\section{Conclusion \& Future Work} \label{sec:conclusion}
\begin{table}[h!]   
    \centering \small
    \caption{Final results}
    \begin{tabular}{ |p{0.03\linewidth} |p{0.4\linewidth} |c | c | c | c | c|}  \hline
    & \textbf{Type} & \textbf{Accuracy} & \textbf{Precision} & \textbf{Recall} & \textbf{F1} & \textbf{MCC} \\ \hline
    \multirow{2}{*}{\rotatebox{90}{\textbf{EE}}} & 
    1. Entity Mention Detection (EMD) + Event Detection (ED) & 0.977 & 0.933 & 0.949 & 0.941 & - \\ \cline{2-7}
    & 2. Argument Role Prediction (ARP) & 0.914 & 0.913 & 0.914 & 0.913 & 0.840 \\ \hline
    \multirow{3}{*}{\rotatebox{90}{\textbf{Prop.}}} & 
    1. Event Polarity & 0.821 & 0.697 & 0.917 & 0.792 & 0.717 \\ \cline{2-7}
    & 2. Event Modality & 0.852 & 0.803 & 0.842 & 0.822 & 0.771 \\ \cline{2-7}
    & 3. Event Intensity & 0.702 & 0.765 & 0.664 & 0.751 & 0.665 \\ \hline
    \end{tabular}
    \label{table:final_results}
\end{table}

It is known that training models via the traditional approach of supervised learning requires a substantial amount of annotated data. This challenge becomes even more apparent for a lower-resource domain such as Finance and Economics. To produce accurate event extraction models from the \textit{CrudeOilNews} corpus, we leveraged the effectiveness of transfer learning to build event extraction and event classification models with the best possible performance despite limited resources. 

%Our solution leverages the effectiveness of Transfer Learning to build event extraction and event classification models with the best possible performance despite the limited resources. 
In the transfer learning setting, we are able to utilize source task $T_s$ or source domain $D_s$ to improve model embeddings and representations in order to produce better model performance in target task $T_t$ or target domain $D_t$. Based on the experiment results, we come to the following conclusions:
\begin{enumerate}
    \item Domain Adaptive Pre-Training: Further pre-training BERT on in-domain data to produce ComBERT produced better contextualized embedding that improved model performance; 
    \item Combination of Multi-Task Learning and Sequential Transfer Learning: After experimenting with various permutations of MTL and STL sub-tasks configurations, the best configuration for event extraction in \textit{CrudeOilNews} is EMD + ED  $\rightarrow$ ARP, where EMD + ED are trained via MTL before sequentially transferring the resulting model to train on ARP. 
    \item Cross-domain Sequential Transfer Learning: Source datasets annotated for the same tasks can be used to boost the performance of event property classification in \textit{CrudeOilNews} corpus despite the severe class imbalance. However, it is worth noting that selecting a source domain with high source-target similarity is vital to have the best model performance and to avoid the undesirable result of negative transfer. 
\end{enumerate}
The final model performance for each sub-task is tabulated and shown again in Table \ref{table:final_results} for easier reference. Even though we have managed to use transfer learning to achieve better model performance, one of the major weaknesses of the framework is that the current event ontologies are exhaustive and will not cater to new events. As part of future work, we are interested in investigating ways to increase event coverage for crude oil news. One promising direction is to use zero-shot learning like in \cite{huang-etal-2018-zero}. Another promising direction is to investigate domain adaptation of event extraction from crude oil news to a similar domain, such as in gold-related news or even FoRex (Foreign Exchange) news. 

%
%A straight forward solution is to first automatically produce more training data, and then use mixture data containing both original gold data and newly generated ones for model training. Many methods were explored on how to expand a small set of labeled data to larger corpus. These methods come in different names such as semi-supervision, weak supervision, distant supervision, unsupervised learning and etc, with the aim to grow the corpus to include more types of events or leveraging external resources such as FrameNet, Abstract Meaning Representation (AMR), and etc. 

\clearpage
\appendix

\section{Target Dataset: CrudeOilNews Corpus} \label{app_dataset}
\begin{itemize}
    \item Tasks: Event Extraction, Event Modality, Polarity and Intensity Classification.
    \item Domain: Commodity news surrounding Crude Oil. 
    \item Size: 425 documents, 7,059 sentences, 10,578 events,  and 22,267 event arguments.
    \item Event Types: Movement-down-loss, movement-up-gain, movement-flat, cause-movement-down-loss, caused-movement-up-gain, position-high, position-low, slow-weak, grow-strong, prohibiting, oversupply, shortage, civil-unrest, embargo, geo-political tension, crisis and negative-sentiment.
\end{itemize} 

\begin{figure*}[ht!]
\includegraphics[width=0.9\textwidth]{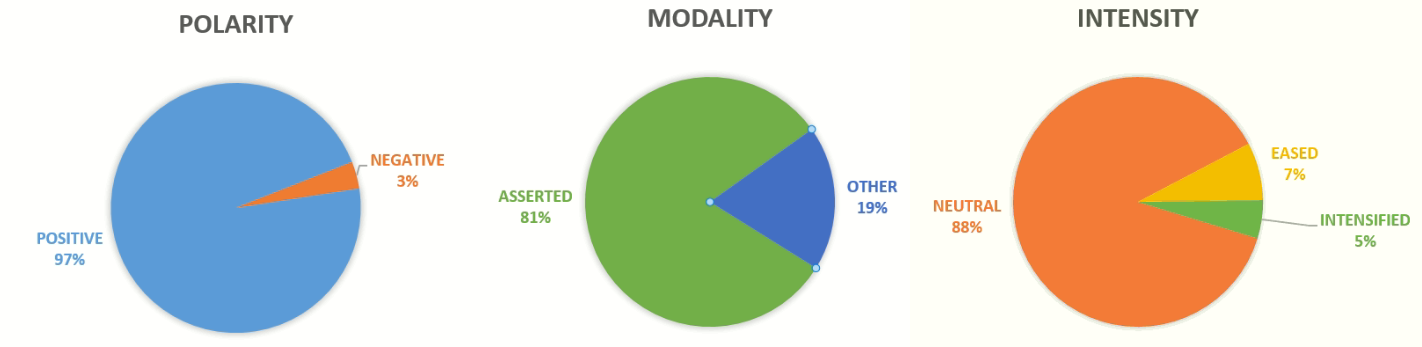}
\caption{Event Polarity, Modality and Intensity Distribution}
\label{fig:class_imbalance}
\end{figure*}

Here is a list of characteristics exhibited by this dataset and each pose a unique challenge to the overall event extraction task:
\begin{enumerate}
    \item Limited amount of labeled data;
    \item Class imbalance / topic bias: serious class imbalance in event Properties distribution as shown in Figure \ref{fig:class_imbalance} where the majority class outnumbers the minority classes by a large margin;
    %\item Multiple events existing in the same sentence, making event extraction more challenging than one-event-one-sentence cases. This is equally true for event properties classification. Even though Polarity / Modality / Intensity cue words appear in a sentence, it is important to know which event these cue words' scope cover;
    \item Homogenous entity types but playing different argument roles; \textbf{One or two examples here will be good.}
    \item Number intensity: Numbers (e.g., price, difference, percentage of change) and dates (including date of the opening price, dates of closing price) are abundant in commodity news.
\end{enumerate}

\subsection{Event Properties Examples}
\begin{enumerate}
    \item Event Polarity
        \begin{itemize}
            \item {\fontfamily{qcr}\selectfont POSITIVE}: \textit{OPEC members agreed to \underline{cut} oil supplies.}
            \item {\fontfamily{qcr}\selectfont NEGATIVE}: \textit{OPEC countries \textbf{refused} to \underline{cut} oil suppliers.}
        \end{itemize}
    \item Event Modality:
        \begin{itemize}
            \item {\fontfamily{qcr}\selectfont ASSERTED}: \textit{Saudi Arabia continues to \underline{cut} its production.}
            \item {\fontfamily{qcr}\selectfont OTHER}: \textit{Analysts were \textbf{anticipating} oil inventories to \underline{fall} by 800,000.} 
        \end{itemize}
    \item Event Intensity:
        \begin{itemize}
            \item {\fontfamily{qcr}\selectfont NEUTRAL}: \textit{Oil \underline{rise} third day in a row to all time high.}
            \item {\fontfamily{qcr}\selectfont EASED}: \textit{Libya 's civil \underline{strife} has been \textbf{eased} by potential peace talks.}
            \item {\fontfamily{qcr}\selectfont INTENSIFIED}: \textit{...could hit Iraq 's output and \textbf{deepen} a supply \underline{shortfall}.}
        \end{itemize}
\end{enumerate}

\section{Source Datasets} 

\begin{enumerate}
    \item {\textbf{ACE2005}} 
        \begin{itemize}
            \item Tasks: Event Extraction, Event Modality and Polarity classification.
            \item Domain: General, corpus made up of conversations, broadcast news, newsgroups, weblogs.
            \item Size: 587 documents, 5,055 events, 18,927 sentences, 6,040 event arguments, 34,474 entities.
            \item Event Types: Life, Movement, Transaction, Business, Conflict, Contact, Personnel, Justice    
        \end{itemize}    
    \item \textbf{SENTiVENT} \cite{jacobs2021sentivent}
        \begin{itemize}
            \item Tasks: Event Extraction, Event Modality and Polarity Classification.
            \item Domain: Business / Finance News
            \item Size: 288 documents, 6,203 events, 6,883 sentences, 13,675 arguments.
            \item Event Types: CSR / Brand, Deal, Dividend, Employment, Expense, Facility, Financial Report, Financing, Investing, Legal, Macroeconomics, Merger / Acquisition, Product / Service, Profit / Loss, Rating, Revenue, Sales Volume, Security Value
        \end{itemize} 
    \item \textbf{ConanDoyle} \cite{morante2012sem}
        \begin{itemize}
            \item Tasks: Negation detection
            \item Domain: Fiction / stories
            \item Size: 3640 sentences from The Hound of the Baskervilles story, out of which 850 contain negations, and 783 sentences from The Adventure of Wisteria Lodge story, out of which 145 contain negations.
            \item Description: a corpus of Conan Doyle stories annotated with negation cues and their scopes, as well as the event or property that is negated. Negation cues are made up of the following types: lexical, syntactic, and morphological.
        \end{itemize}
    \item \textbf{SOCC(neg)} \cite{kolhatkar2020sfu}
        \begin{itemize}
            \item Tasks: Negation detection
            \item Domain: opinion, review and comments
            \item Size: 10399 opinion articles (editorials, columns and op-eds); 663,173 comments from 303,665 comment threads, from the main Canadian daily newspaper (from January 2012 to December 2016)
            \item Description: The corpus is organized into three subcorpora: the articles corpus, the comments corpus, and the comment-threads corpus. Only the articles sub-corpora is used for this work.
        \end{itemize}
    \item \textbf{10kFinStatement(unc)} \cite{theil2018word}
        \begin{itemize}
            \item Tasks: Uncertainty detection
            \item Domain: Finance - 10-k reports
            \item Size: 1000 sentences from 10-Ks\footnote{A 10-K is a comprehensive report filed annually by public companies about their financial performance.}
            \item Description: each labeled with \textit{certain} and \textit{uncertain} but does not have uncertainty cue words nor scope annotated. 
        \end{itemize}
    \item \textbf{Wikipedia-CoNLL2010(unc)} 
        \begin{itemize}
            \item Tasks: dataset is provided as part of CoNLL2010 shared task - Learning to Detect Hedges and their Scope in Natural Language Text.
            \item Domain: General - Wikipedia articles
            \item Size: 2,186 paragraphs collected from Wikipedia archives were also offered as Task1 training data (11,111 sentences containing 2,484 uncertain ones). The evaluation dataset contained 2,346 Wikipedia paragraphs with 9,634 sentences, out of which 2234 were uncertain.
            \item Description: 
        \end{itemize}
    \item \textbf{Reviews(neg \& unc)} \cite{konstantinova2012review}
        \begin{itemize}
            \item Tasks: Negation and Uncertainty detection
            \item Domain: Reviews of movie, book and consumer product, taken from \href{www.Epinions.com}{www.Epinions.com}.
            \item Size: 400 documents 
        \end{itemize}
\end{enumerate}

\section{Additional results}  \label{app:additional_results}
Detail breakdown of Event Extraction results by their respective classes are presented here. Entity Mention Detection (EMD) + Event Detection (ED) results are presented in Table \ref{table:entity_trigger_results} while Argument Role Prediction (ARP) results are listed in Table \ref{table:ARP_results}.

\begin{table}[h!]   
    \centering \small
    \caption{Entity Mention + Event Trigger Extraction}
    \begin{tabular}{ |p{0.03\linewidth} |p{0.50\linewidth} | c | c | c|}  \hline
    & \textbf{Type} & \textbf{P} & \textbf{R} & \textbf{F1} \\ \hline
    \multirow{21}{*}{\rotatebox{90}{Entity Mention}} 
    & 1. {\fontfamily{qcr}\selectfont COMMODITY} & 0.98 & 0.98 & 0.98 \\ \cline{2-5}
    & 2. {\fontfamily{qcr}\selectfont COUNTRY} & 0.97 & 0.98 & 0.98 \\ \cline{2-5}
    & 3. {\fontfamily{qcr}\selectfont DATE} & 0.94 & 0.96 & 0.95\\ \cline{2-5}
    & 4. {\fontfamily{qcr}\selectfont DURATION} & 0.87 & 0.90 & 0.89\\ \cline{2-5}
    & 5. {\fontfamily{qcr}\selectfont ECONOMIC\_ITEM} & 0.89 & 0.95 & 0.92\\ \cline{2-5}
    & 6. {\fontfamily{qcr}\selectfont FINANCIAL\_ATTRIBUTE} & 0.98 & 0.99 & 0.99\\ \cline{2-5}
    & 7. {\fontfamily{qcr}\selectfont FORECAST\_TARGET} & 0.92 & 0.95 & 0.94\\ \cline{2-5}
    & 8. {\fontfamily{qcr}\selectfont GROUP} & 0.56 & 0.50 & 0.53\\ \cline{2-5}
    & 9. {\fontfamily{qcr}\selectfont LOCATION} & 0.90 & 0.91 & 0.90\\ \cline{2-5}
    & 10. {\fontfamily{qcr}\selectfont MONEY} & 0.92 & 0.94 & 0.93\\ \cline{2-5}
    & 11. {\fontfamily{qcr}\selectfont NATIONALITY} & 0.96 & 0.92 & 0.94\\ \cline{2-5}
    & 12. {\fontfamily{qcr}\selectfont NUMBER} & 0.97 & 0.88 & 0.92\\ \cline{2-5}
    & 13. {\fontfamily{qcr}\selectfont ORGANIZATION} & 0.94 & 0.96 & 0.95\\ \cline{2-5}
    & 14. {\fontfamily{qcr}\selectfont OTHER\_ACTIVITY} & 0.33 & 0.50 & 0.40\\ \cline{2-5}
    & 15. {\fontfamily{qcr}\selectfont PERCENT} & 0.96 & 0.95 & 0.95 \\ \cline{2-5}
    & 16. {\fontfamily{qcr}\selectfont PERSON} & 0.98 & 0.97 & 0.98 \\ \cline{2-5}
    & 17. {\fontfamily{qcr}\selectfont PHENOMENON} & 0.0 & 0.0 & 0.0\\ \cline{2-5}
    & 18. {\fontfamily{qcr}\selectfont PRICE\_UNIT} & 0.96 & 0.99 & 0.97\\ \cline{2-5}
    & 19. {\fontfamily{qcr}\selectfont PRODUCTION\_UNIT} & 0.90 & 0.95 & 0.92\\ \cline{2-5}
    & 20. {\fontfamily{qcr}\selectfont QUANTITY} & 0.80 & 0.92 & 0.86\\ \cline{2-5}
    & 21. {\fontfamily{qcr}\selectfont STATE\_OR\_PROVINCE} & 0.77 & 0.85 & 0.80\\ \hline
    \multirow{18}{*}{\rotatebox{90}{Event Trigger}} 
    & 1. {\fontfamily{qcr}\selectfont CAUSE\_MOVEMENT\_DOWN\_LOSS} & 0.92 & 0.93 & 0.92  \\ \cline{2-5}
    & 2. {\fontfamily{qcr}\selectfont CAUSE\_MOVEMENT\_UP\_GAIN} & 0.87 & 0.89 & 0.88\\ \cline{2-5}
    & 3. {\fontfamily{qcr}\selectfont CIVIL\_UNREST} & 1.00 & 0.92 & 0.96\\ \cline{2-5}
    & 4. {\fontfamily{qcr}\selectfont CRISIS} & 1.00 & 1.00 & 1.00\\ \cline{2-5}
    & 5. {\fontfamily{qcr}\selectfont EMBARGO} & 0.97 & 1.00 & 0.98\\ \cline{2-5}
    & 6. {\fontfamily{qcr}\selectfont GEOPOLITICAL\_TENSION} & 0.75 & 0.88 & 0.81\\ \cline{2-5}
    & 7. {\fontfamily{qcr}\selectfont GROW\_STRONG} & 0.79 & 0.84 & 0.81\\ \cline{2-5}
    & 8. {\fontfamily{qcr}\selectfont MOVEMENT\_DOWN\_LOSS} & xx & xx & xx\\ \cline{2-5}
    & 9. {\fontfamily{qcr}\selectfont MOVEMENT\_FLAT} & 1.00 & 0.86 & 0.92\\ \cline{2-5}
    & 10. {\fontfamily{qcr}\selectfont MOVEMENT\_UP\_GAIN} & xx & xx & xx\\ \cline{2-5}
    & 11. {\fontfamily{qcr}\selectfont NEGATIVE\_SENTIMENT} & 0.93 & 0.94 & 0.93\\ \cline{2-5}
    & 12. {\fontfamily{qcr}\selectfont OVERSUPPLY} & 0.80 & 0.83 & 0.82\\ \cline{2-5}
    & 13. {\fontfamily{qcr}\selectfont POSITION\_HIGH} & 0.91 & 1.00 & 0.96\\ \cline{2-5}
    & 14. {\fontfamily{qcr}\selectfont POSITION\_LOW} & 0.91 & 0.97 & 0.94\\ \cline{2-5}
    & 15. {\fontfamily{qcr}\selectfont PROHIBITING} & 0.83 & 0.83 & 0.83\\ \cline{2-5}
    & 16. {\fontfamily{qcr}\selectfont SHORTAGE} & 0.91 & 1.00 & 0.95\\ \cline{2-5}
    & 17. {\fontfamily{qcr}\selectfont SLOW\_WEAK} & 0.93 & 0.73 & 0.82\\ \cline{2-5}
    & 18. {\fontfamily{qcr}\selectfont TRADE\_TENSIONS} & 0.75 & 0.88 & 0.81\\ \hline
    \end{tabular}
    \label{table:entity_trigger_results}
\end{table}

\begin{table}[h!]   
    \centering \small
    \caption{Argument Role Prediction}
    \begin{tabular}{ |p{0.6\linewidth} | c | c| c| }  \hline
    \textbf{Type} & \textbf{P} & \textbf{R} & \textbf{F1} \\ \hline
    1. {\fontfamily{qcr}\selectfont NONE} & 0.94 & 0.95 & 0.95 \\ \hline
    2. {\fontfamily{qcr}\selectfont ATTRIBUTE} & 0.92 & 0.93 & 0.93 \\\hline
    3. {\fontfamily{qcr}\selectfont ITEM} & 0.88 & 0.84 & 0.86 \\ \hline
    4. {\fontfamily{qcr}\selectfont FINAL\_VALUE} & 0.85 & 0.78 & 0.82 \\ \hline
    5. {\fontfamily{qcr}\selectfont INITIAL\_VALUE} & 0.50 & 0.30 & 0.37 \\ \hline
    6. {\fontfamily{qcr}\selectfont DIFFERENCE} & 0.89 & 0.91 & 0.90 \\ \hline
    7. {\fontfamily{qcr}\selectfont REFERENCE\_POINT} & 0.79 & 0.77 & 0.78 \\ \hline
    8. {\fontfamily{qcr}\selectfont INITIAL\_REFERENCE\_POINT} & 0.43 & 0.53 & 0.47 \\ \hline
    9. {\fontfamily{qcr}\selectfont CONTRACT\_DATE} & 1.00 & 0.94 & 0.97 \\ \hline
    10. {\fontfamily{qcr}\selectfont DURATION} & 0.83 & 0.87 & 0.85 \\ \hline
    11. {\fontfamily{qcr}\selectfont TYPE} & 0.79 & 0.74 & 0.76 \\ \hline
    12. {\fontfamily{qcr}\selectfont IMPOSER} & 0.84 & 1.00 & 0.91\\ \hline
    13. {\fontfamily{qcr}\selectfont IMPOSEE} & 0.84 & 0.84 & 0.84 \\ \hline
    14. {\fontfamily{qcr}\selectfont PLACE} & 0.81 & 0.77 & 0.79 \\ \hline
    15. {\fontfamily{qcr}\selectfont SUPPLIER\_CONSUMER} & 0.84 & 0.82 & 0.83 \\ \hline
    16. {\fontfamily{qcr}\selectfont IMPACTED\_COUNTRIES} & 0.93 & 0.87 & 0.90 \\ \hline
    17. {\fontfamily{qcr}\selectfont PARTICIPATING\_COUNTRIES} & 0.86 & 0.92 & 0.89 \\ \hline
    18. {\fontfamily{qcr}\selectfont FORECASTER} & 0.80 & 0.63 & 0.71 \\ \hline
    19. {\fontfamily{qcr}\selectfont FORECAST} & 0.76 & 0.59 & 0.67 \\ \hline
    20. {\fontfamily{qcr}\selectfont SITUATION} & 0.67 & 0.50 & 0.57 \\ \hline
    \end{tabular}
    \label{table:ARP_results}
\end{table}

%List of events found in ACE2005:
%1. Life - Be-Born
%2. Life - Marry
%3 Life - Divorce
%4 Life - Injure
%5 Life - Die 
%6 Movement - Transport
%7 Transaction - Transfer Ownership
%8 Transaction - Transfer Money
%9 Business - Start-Org
%10 Business - Merge-Ord
%11 Business - Declare-Bankruptcy
%12 Business - End-Org
%13 Conflict - Attack
%14 Conflict - Demonstrate
%15 Contact - Meet
%16 Contact - Phone-write
%17 Personnel - start-position
%18 Personnel - End-Position
%19 Personnel - Nominate
%20 Personnel - Elect
%21 Justice - Arrest-Jail
%22 Justice - Release-Paraole
%23 Justice - Charge-indict
%24 Justice - Sue
%25 Justice - Convict
%26 Justice - Sentence
%27 Justice - Fine
%28 Justice - Execute
%29 Justice - Extradite
%30 Justice - Acquit
%31 Justice - Appeal
%32 Justice - Pardon

%CSR / Brand
%deal
%Dividend
%Employment
%Expense
%Facility
%Financial Report
%Financing
%Investing
%Legal
%Macroeconomics
%Merger / Acquisition
%Product / Service
%Profit / Loss
%Rating
%Revenue
%Sales Volume
%Security Value

%Macro-economic events are aggregated indicators suchas sectorial trends, policy, GDP, unemployment rates,national income, price indices, and the interrelationsamong the different markets and sectors of the economy.Includes events, speculation, expectation, forecasts, an-nouncements, changes on a company’s position in themarket, the state of the market, sector or country ingeneral, headwinds, tailwinds, business trends, marketshare, consumer spending, etc. We include discussionson a company’s position in the market (e.g. compared to competitors).

\bibliography{sample}

\end{document}